\documentclass[sn-mathphys,Numbered]{sn-jnl}

\usepackage{graphicx}%
\usepackage{multirow}%
\usepackage{amsmath,amssymb,amsfonts}%
\usepackage{amsthm}%
\usepackage{mathrsfs}%
\usepackage[title]{appendix}%
\usepackage{xcolor}%
\usepackage{textcomp}%
\usepackage{manyfoot}%
\usepackage{booktabs}%
\usepackage{algorithm}%
\usepackage{algorithmicx}%
\usepackage{algpseudocode}%
\usepackage{listings}%

\usepackage{makecell}
\usepackage{pifont}
\usepackage{placeins}
\newcommand{\cmark}{\ding{51}}%
\usepackage{soul}%

\theoremstyle{thmstyleone}%
\theoremstyle{thmstyletwo}%

\theoremstyle{thmstylethree}%

\begin{document}

\title{Deep Learning for Computer Vision based Activity Recognition and Fall Detection of the Elderly: a Systematic Review}


\author*[1,2,3]{\fnm{F. Xavier} \sur{Gaya-Morey}}\email{francesc-xavier.gaya@uib.es}
\equalcont{These authors contributed equally to this work.}

\author[1,2,3]{\fnm{Cristina} \sur{Manresa-Yee}}\email{cristina.manresa@uib.es}
\equalcont{These authors contributed equally to this work.}

\author[1,2,3]{\fnm{José M.} \sur{Buades-Rubio}}\email{josemaria.buades@uib.es}\equalcont{These authors contributed equally to this work.}

\affil[1]{\orgdiv{Group of Computer Graphics, Computer Vision and AI (UGIVIA)}, \orgname{Universitat de les Illes Balears (UIB)}, \orgaddress{\street{Carretera de Valldemossa, km 7.5}, \city{Palma}, \postcode{07122}, \state{Illes Balears}, \country{Spain}}}

\affil[2]{\orgdiv{Research Institute of Health Sciences (IUNICS)}, \orgname{Universitat de les Illes Balears (UIB)}, \orgaddress{\street{Carretera de Valldemossa, km 7.5}, \city{Palma}, \postcode{07122}, \state{Illes Balears}, \country{Spain}}}

\affil[3]{\orgdiv{Department of Mathematics and Computer Science}, \orgname{Universitat de les Illes Balears (UIB)}, \orgaddress{\street{Carretera de Valldemossa, km 7.5}, \city{Palma}, \postcode{07122}, \state{Illes Balears}, \country{Spain}}}



\abstract{As the proportion of elderly individuals in developed countries continues to rise globally, addressing their healthcare needs, particularly in preserving their autonomy, is of paramount concern. A growing body of research focuses on Ambient Assisted Living (AAL) systems, aimed at alleviating concerns related to the independent living of the elderly. This systematic review examines the literature pertaining to fall detection and Human Activity Recognition (HAR) for the elderly, two critical tasks for ensuring their safety when living alone. Specifically, this review emphasizes the utilization of Deep Learning (DL) approaches on computer vision data, reflecting current trends in the field. A comprehensive search yielded 2,616 works from five distinct sources, spanning the years 2019 to 2023 (inclusive). From this pool, 151 relevant works were selected for detailed analysis. The review scrutinizes the employed DL models, datasets, and hardware configurations, with particular emphasis on aspects such as privacy preservation and real-world deployment. The main contribution of this study lies in the synthesis of recent advancements in DL-based fall detection and HAR for the elderly, providing insights into the state-of-the-art techniques and identifying areas for further improvement. Given the increasing importance of AAL systems in enhancing the quality of life for the elderly, this review serves as a valuable resource for researchers, practitioners, and policymakers involved in developing and implementing such technologies.}

\keywords{Human activity recognition, Fall detection, Ambient assisted living, Deep learning, Computer vision, Elderly}



\maketitle

\section{Introduction}




    The global population is experiencing rapid growth, accompanied by a significant increase in life expectancy, particularly in developed countries. Bloom and Luca \cite{Bloom2016} note that life expectancy in China and India has surged by nearly 30 years since 1950. Consequently, a substantial portion of the population in developed nations, approximately 20\%, is aged 60 and above, a figure projected to surpass 30\% in the next four decades.


    With this demographic shift comes a growing concern for elderly\footnote{In this work, the term "elderly" refers to individuals over 65 years old, as commonly used in the field of Computer Science. However, terminology may vary in other fields; for instance, the term "older adults" is often preferred in Psychology.} care, as the need for assistance and support rises proportionately. Among the myriad challenges faced by the elderly, falls represent a particularly prevalent and perilous occurrence. The World Health Organization highlights alarming statistics on falls, identifying them as the second leading cause of unintentional injury deaths worldwide. Each year, an estimated 684,000 individuals succumb to fall-related injuries globally, with an additional 37.3 million falls severe enough to necessitate medical attention \cite{factSheetsFalls}. Apart from the physical harm incurred by the elderly, the economic ramifications are substantial, with fall-related treatment costs comprising a significant portion of healthcare expenditures in various countries such as the USA, Australia, EU15 and the United Kingdom \cite{Heinrich2010}.


    Automated fall detection for the elderly is feasible through data collected from wearable or environmental devices, such as accelerometers, gyroscopes, and cameras. Furthermore, Human Activity Recognition (HAR) holds promise for diverse applications, ranging from automatic life-logging to identifying patterns indicative of illness \cite{climent_pérez2020review, Khodabandehloo2021}. Vision data from cameras is increasingly utilized for fall detection and HAR tasks due to its numerous advantages over wearable devices or other sensors. These advantages include the ability to detect multiple events simultaneously, suitability for various subjects, environments, and tasks, as well as ease of installation and visual verification of data \cite{Nizam2020}.


    From an algorithmic standpoint, Deep Learning (DL) has revolutionized digital image processing, emerging as the state-of-the-art approach in numerous domains \cite{DLvsTCV}. Over recent years, a plethora of DL architectures have been developed and evaluated for computer vision tasks, prompting the exploration of suitable models for HAR and fall detection among older adults.


    In this study, we conduct a Systematic Literature Review (SLR) focusing on DL-based HAR and fall detection using vision data for elderly care. Our review strictly adheres to the guidelines outlined for conducting SLRs in Software Engineering by Kitchenham and Charters \cite{slr-software}, providing a structured methodology and rigorous analysis. The document is structured as follows: we first delve into the background of the study, encompassing previous reviews and defining key concepts; we then enumerate the review questions; next, we elaborate on the review methods, detailing data sources, search strategy, study selection, quality assessment, and data extraction; subsequently, we analyze the resulting studies comprehensively to address the review questions; the discussion section synthesizes our findings and addresses the review questions; finally, we present the conclusions derived from the SLR.

\section{Background}


    In accordance with the guidelines provided by Kitchenham and Charters \cite{slr-software}, it is imperative to summarize previous reviews prior to conducting the SLR, thereby substantiating its necessity. Hence, we briefly outline related reviews and surveys from the past three years, as older reviews cannot encompass the most recent studies. The full list can be found in Table \ref{tab:review-comparison}, providing a visual comparison of the main disparities.

    Guerra et al. \cite{guerra2023ambient} studied the current state-of-the-art of Ambient Assisted Living (AAL) for frail individuals, including the elderly and disabled, encompassing both wearable and non-wearable solutions. They explored common steps in the Human Activity Recognition (HAR) processing chain. Similarly, Kumar et al. \cite{kumar2023survey} delved into various types of data used for HAR, elucidating common datasets, approaches, and challenges, but not including the elderly population or fall detection. In \cite{tay2023review}, Tay et al. investigated abnormal behavior detection, such as fall detection, repetition of activities, and accidents. They explored multiple solutions, including visual and wearable sensors, and both conventional and Deep Learning (DL) approaches. A review by Momin et al. \cite{Momin2022} explores activity pattern monitoring using depth sensors, considering this visual data as a privacy-preserving alternative to RGB video or images for older adults. The studies are categorized based on the computing technique utilized and the datasets used are analyzed. Olugbade et al. \cite{Olugbade2022} conducted a scoping review on datasets utilized for HAR and fall detection, resulting in an extensive compilation of over 700 datasets of various modalities. Multiple taxonomies were developed to categorize the datasets by population groups, data types, and creation purposes, among others, although only four datasets include elderly subjects. Alam et al. \cite{alam2022vision} conducted a review specifically on DL-based fall detection systems, analyzing different fall types, popular datasets, evaluation metrics, and architectural variations. Rastogi et al. \cite{rastogi2022human} reviewed a broader range of tasks, including falls and other relevant information extracted from video sequences, such as body shape changes, posture, and gait. Another relevant review is by Gutiérrez et al. \cite{gutiérrez2021comprehensive}, also centered on fall detection, which describes common processing steps, ML models, datasets, metrics, and tracking techniques, with most studies utilizing RGB and depth data.

\begin{table}[h]
    \caption{Comparison of previous reviews with ours. By columns, important aspects taken into account in this review, and whether they are addressed or not by each review. From left to right: if the review is systematic; focuses on the exploration of DL solutions; targets elderly people as the users of the system; centers on the use of vision data; explores the Fall Detection and the Human Activity Recognition tasks; explores the use of RGB, depth or infrared data; takes privacy as a critical concern; describes the hardware used in the found studies; and if it studies the deployment of the FD and HAR systems in real environments.}
    \label{tab:review-comparison}
    \setlength{\tabcolsep}{3pt}\begin{tabular*}{1.\textwidth}{c|c|c|ccc|cc|ccc|ccc}
         \multicolumn{1}{l|}{\multirow{2}{*}{Work}} & \multicolumn{1}{l|}{\multirow{2}{*}{Year}} & \multicolumn{1}{l|}{\multirow{2}{*}{System.}} & \multicolumn{3}{c|}{Focus} & \multicolumn{2}{c|}{Task} & \multicolumn{3}{c|}{Data} & \multicolumn{3}{c}{AAL} \\
         & & \multicolumn{1}{c|}{} & DL & Elderly & \multicolumn{1}{c|}{CV} & FD & \multicolumn{1}{c|}{HAR} & RGB & Depth & \multicolumn{1}{c|}{IR} & Privacy & HW& Deployment \\ \midrule
        Ours & 2024                       & \cmark                      & \cmark & \cmark  & \cmark & \cmark      & \cmark     & \cmark & \cmark & \cmark & \cmark  & \cmark   & \cmark    \\
        \cite{guerra2023ambient} & 2023          &                             &        &  \cmark &        & \cmark      & \cmark     & \cmark & \cmark & \cmark & \cmark  &          &            \\
        \cite{kumar2023survey} & 2023            &                             &        &         &        &       & \cmark     & \cmark & \cmark &        &         &          &            \\
        \cite{tay2023review} & 2023              &                             &        & \cmark  &        & \cmark      &            & \cmark & \cmark & \cmark &         &          &            \\
        \cite{Momin2022} & 2022                  &                             &        & \cmark  & \cmark & \cmark      & \cmark     &        & \cmark &        & \cmark  &          &            \\
        \cite{Olugbade2022} & 2022               & \cmark                      &        &         &        & \cmark      & \cmark     & \cmark & \cmark & \cmark &         &          &            \\
        \cite{alam2022vision} & 2022             &                             & \cmark & \cmark  & \cmark & \cmark      &            & \cmark & \cmark & \cmark & \cmark  &          &            \\
        \cite{rastogi2022human} & 2022           &                             &        & \cmark  & \cmark & \cmark      & \cmark     & \cmark & \cmark & \cmark &         &          &            \\
        \cite{gutiérrez2021comprehensive} & 2021 &                             &        & \cmark  & \cmark & \cmark      &            & \cmark & \cmark & \cmark &         &          &            \\
        \bottomrule
    \end{tabular*}
\end{table}


    While prior reviews have addressed various aspects of our research domain, notable differences underscore the necessity of our study. Table \ref{tab:review-comparison} sheds light upon this by displaying the pivotal aspects considered in the current review, along with whether they are addressed or not in the aforementioned reviews.
    

    The sole review exclusively focusing on DL techniques was conducted by Alam et al. \cite{alam2022vision}, which, however, omitted HAR from its scope, thus neglecting a significant portion of studies included in our analysis. In contrast, other reviews encompassed techniques employing handcrafted features or classical vision approaches, reflecting a broader scope than our exclusive focus on DL-based solutions. Furthermore, previous reviews often overlooked the importance of studying DL-related nuances, such as the significance of training datasets, architectural considerations, and feature extraction methods. In our review, we meticulously categorize and elucidate these nuances through a comprehensive taxonomy of identified techniques.
    

    Another notable observation is the limited attention given to HAR in several previous reviews, with some omitting the task altogether. As a result, our review unveils a greater number of studies dedicated to fall detection and HAR in the elderly. Additionally, our analysis delves deeper into the intricacies of these tasks, providing a more comprehensive understanding.


    Moreover, only a few reviews explored applications within AAL systems and the associated privacy implications. Hardware specifications, beyond the prevalent use of Kinect cameras, were rarely examined, and the effective deployment of fall detection or HAR systems was not thoroughly explored. In contrast, our review emphasizes these aspects, which are pivotal in facilitating the transference to society.


    Finally, it is worth noting that, apart from \cite{Olugbade2022}, none of the previous reviews adhered to a systematic review process. By rigorously following the systematic review methodology outlined by Kitchenham and Charters \cite{slr-software}, our study ensures a robust and unbiased selection and analysis of relevant studies. We conducted a comprehensive search across various databases, employing well-defined search strings aligned with our research questions. Each study underwent careful quality assessment, and strict exclusion criteria were applied to ensure the inclusion of only the most relevant and high-quality literature. This systematic approach minimizes potential biases and ensures that our review is based on a well-rounded selection of literature.

\section{Review Questions}


    As outlined in \cite{slr-software}, specifying the research questions is a critical aspect of any systematic review, as they guide the entire methodology: from the search process identifying primary studies to address them, to the data extraction process extracting the required data items, and finally to the data analysis synthesizing the data to answer the questions. The review questions for this study are presented in Table \ref{tab:questions}.
    
    \begin{table}[h]
        \caption{Primary and secondary research questions used for this SLR.}
        \label{tab:questions}
        \begin{tabular*}{0.91\textwidth}{ll}
            \toprule
            \textbf{ID} & \textbf{Research Question} \\
            \midrule
            \textbf{RQ1} & 
            \makecell[l]{
                    \textbf{What computer vision deep learning techniques are used for human } \\ 
                    \textbf{activity recognition and fall detection on elderly people?}}\\[0.4cm]
   
            RQ1.1 & 
            What is the preferred data type?\\[0.2cm]
   
            RQ1.2 & 
            What are the most extensively used architectures?\\[0.2cm]
   
            RQ1.3 & 
            What are the most extended datasets?\\[0.2cm]
            
            \midrule
   
            \textbf{RQ2} & 
            \textbf{How can these tasks be deployed successfully in a real environment?}\\[0.2cm]
   
            RQ2.1 & 
            What is the most common hardware (cameras, robots, etc.)?\\[0.2cm]
   
            RQ2.2 & 
            How is privacy of the elderly preserved?\\[0.2cm]
            \bottomrule
            \end{tabular*}
    \end{table}


    The first research question, RQ1, aims to identify the methods used to recognize activities or detect falls among elderly individuals. The choice to specifically investigate HAR and fall detection stemmed from an exploratory initial search, where they emerged as the two most relevant recognition tasks in AAL for the elderly. Given that visual data offers numerous advantages over other sensor data types, such as visual verification and simultaneous subject recognition, and DL has become the state-of-the-art approach in computer vision, conducting an in-depth analysis of the most prevalent methods with these characteristics is crucial for informing future research in this domain. Furthermore, three research subquestions are included regarding common data types (e.g., RGB, depth, thermal, etc.), DL architectures (e.g., CNN, RNN, etc.), and datasets found in the reviewed literature. These subquestions aim to delve deeper into the solution choices at different design steps, which are closely related to various requirements such as privacy preservation, result stability, and inference speed.


    The second research question, RQ2, emerges as a significantly unexplored area, as highlighted in Table \ref{tab:review-comparison} of the Background section. Many previous reviews have focused on the recognition phase of previous studies, enumerating common methods, processing steps, and datasets. However, the effective deployment in real-world scenarios is pivotal for the transfer of such methods to society, and this aspect remains largely unexplored. Works with implementations in real environments, whether through the use of assistive robots or camera-based setups, are expected to be found among the selected studies. Therefore, it is desirable to explore their design choices, setups, and encountered challenges in greater depth. Additionally, privacy is a particularly concerning aspect to consider when dealing with users, especially when utilizing visual data from cameras, and the approaches to addressing it are of interest for future research. For these reasons, RQ2.1 and RQ2.2 delve into common hardware choices and privacy preservation strategies.

\section{Review Methods}


    In this section, we provide a detailed description of the systematic review protocol followed, based on the guidelines outlined by Kitchenham and Charters \cite{slr-software}. Firstly, we list and analyze the primary data sources used, providing visualization of the distribution of studies among these sources. Next, we define the search strategy, which encompasses search terms, synonyms, and time restrictions. Following this, we establish criteria for inclusion and exclusion of studies, followed by the design of a quality assessment checklist to identify and remove low-quality studies. Finally, in the data extraction and synthesis stage, we define how information from each primary study is obtained and outline the specific attributes considered of interest.

    \subsection{Data sources}
    

        For this systematic review, we selected five primary data sources: SCOPUS, Web of Science (WOS), IEEE Xplore Digital Library, ACM Digital Library, and PubMed.
        
        
        SCOPUS and WOS were chosen as comprehensive digital libraries covering a wide range of disciplines, while IEEE Xplore focuses on engineering and technology, ACM Digital Library specializes in computer science, and PubMed is centered on biomedical studies. This selection ensures the inclusion of relevant literature from diverse domains, maximizing the breadth of content considered in our review.

\begin{figure}
             \centering
             \includegraphics[width=0.5\textwidth]{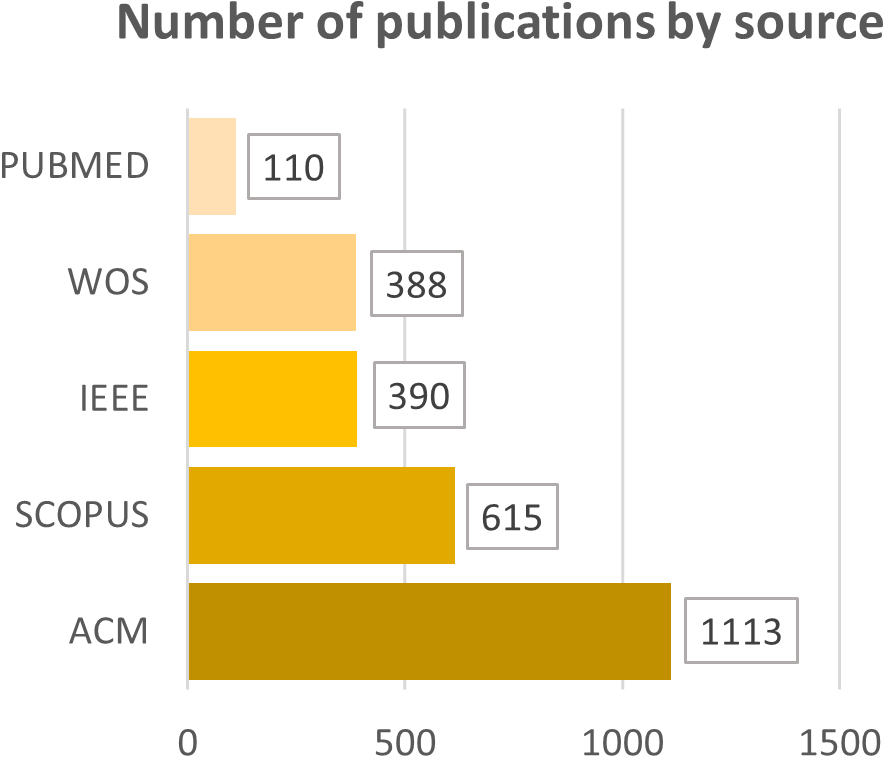}
             \caption{Number of publications obtained from each database, before duplicate removal and study selection (2616 in total).}
            \label{fig:sources}
        \end{figure}

        The distribution of studies retrieved from each source is illustrated in Figure \ref{fig:sources}. As depicted, the majority of studies were sourced from ACM and SCOPUS, with only a small fraction (110 out of a total of 2,616) obtained from PubMed.
        
    \subsection{Search strategy}
    
        


        We constructed different query strings tailored to match the syntax of each digital library while minimizing differences and employing consistent synonyms for the concepts being searched. Each query string connected the various concepts using logical AND, while synonyms for each concept were connected with logical OR. To account for inflection of certain keywords, we utilized the "*" operator after the root word to allow for any possible word endings. In the SCOPUS library, the search was restricted to titles, abstracts, or keywords due to the impracticality of retrieving results otherwise, with the majority being poorly relevant. Conversely, the entire text was searched for in the remaining databases. The primary concepts searched, along with their corresponding lists of synonyms, are as follows:

        \begin{itemize}
        
            \item \textbf{Task to perform (activity recognition or fall detection):} \textit{"action recognition"} OR \textit{"activit* recognition"} OR \textit{"fall* detection"} OR \textit{"behaviour recognition"} OR \textit{"behaviour detection"} OR \textit{"physical activity recognition"}

            \item \textbf{Ambient Assisted Living:} \textit{"monitoring"} OR \textit{"assist* living"} OR \textit{"AAL"} OR \textit{"smart home"} OR \textit{"activit* of daily life"} OR \textit{"activit* of daily living"} OR \textit{"ADL"}

            \item \textbf{Target collective (elderly people):} \textit{"elder*"} OR \textit{"old* people"} OR \textit{"senior"}

            \item \textbf{Kind of data used (Computer Vision):} \textit{"vision"} OR \textit{"rgb"} OR \textit{"video"} OR \textit{"image"} OR \textit{"skeleton"} OR \textit{"depth"} OR \textit{"camera"} OR \textit{"gesture"}
            
        \end{itemize}
        
\begin{figure}
             \centering
             \includegraphics[width=.7\textwidth]{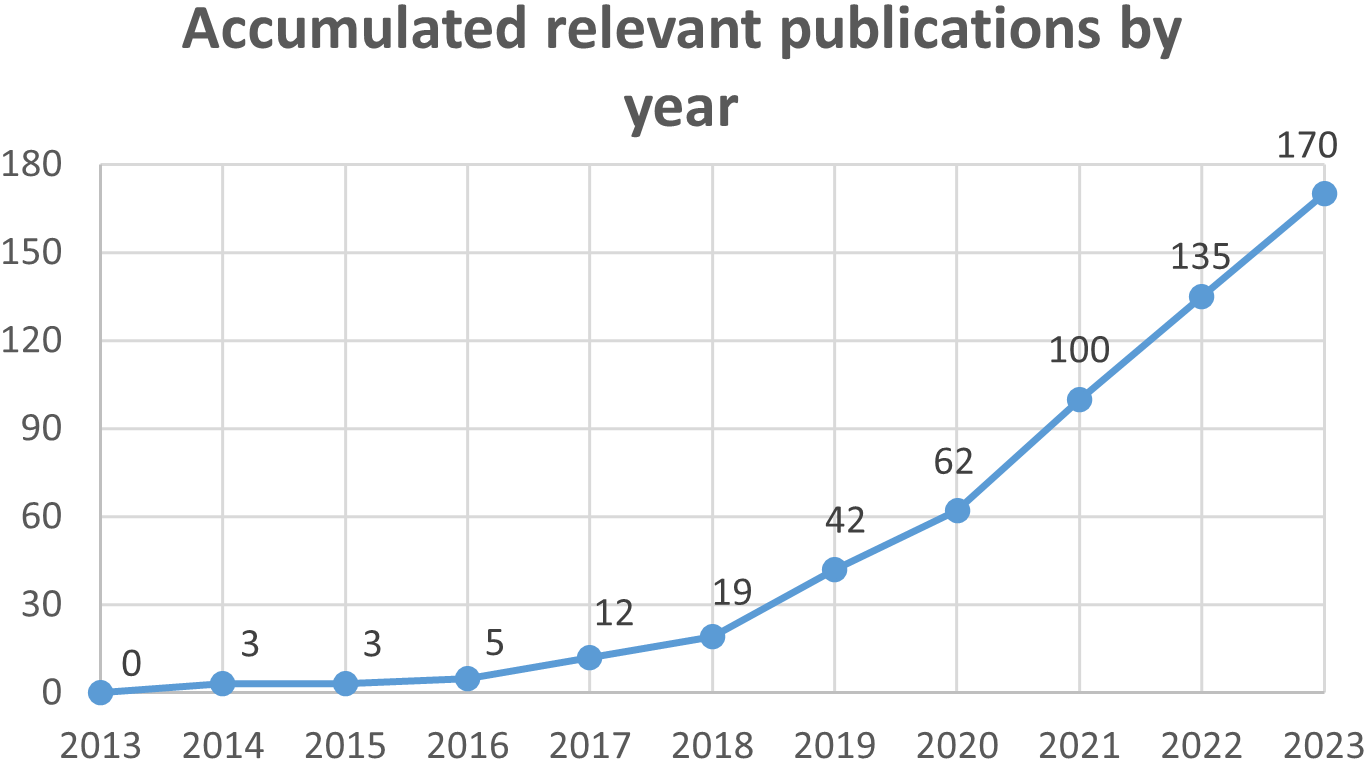}
             \caption{Accumulated publications from 2013 to 2023 (both included) after study selection and quality assessment.}
            \label{fig:years}
        \end{figure}

        Initially, we included studies published from 2013 onwards in the search. However, upon further examination, we observed that the majority of relevant studies were published recently. Consequently, we decided to limit the review to the last five years. Figure \ref{fig:years} displays the accumulated relevant studies from 2013 to 2023. As depicted, only 19 relevant articles were found during the first six years, while 151 were discovered in the last five. This trend underscores the increasing significance of DL-based strategies for HAR and fall detection. By focusing on studies published in the last five years, we aim to gain a deeper analysis of recent trends.
        
\begin{figure}
            \centering
            \includegraphics[width=.6\textwidth]{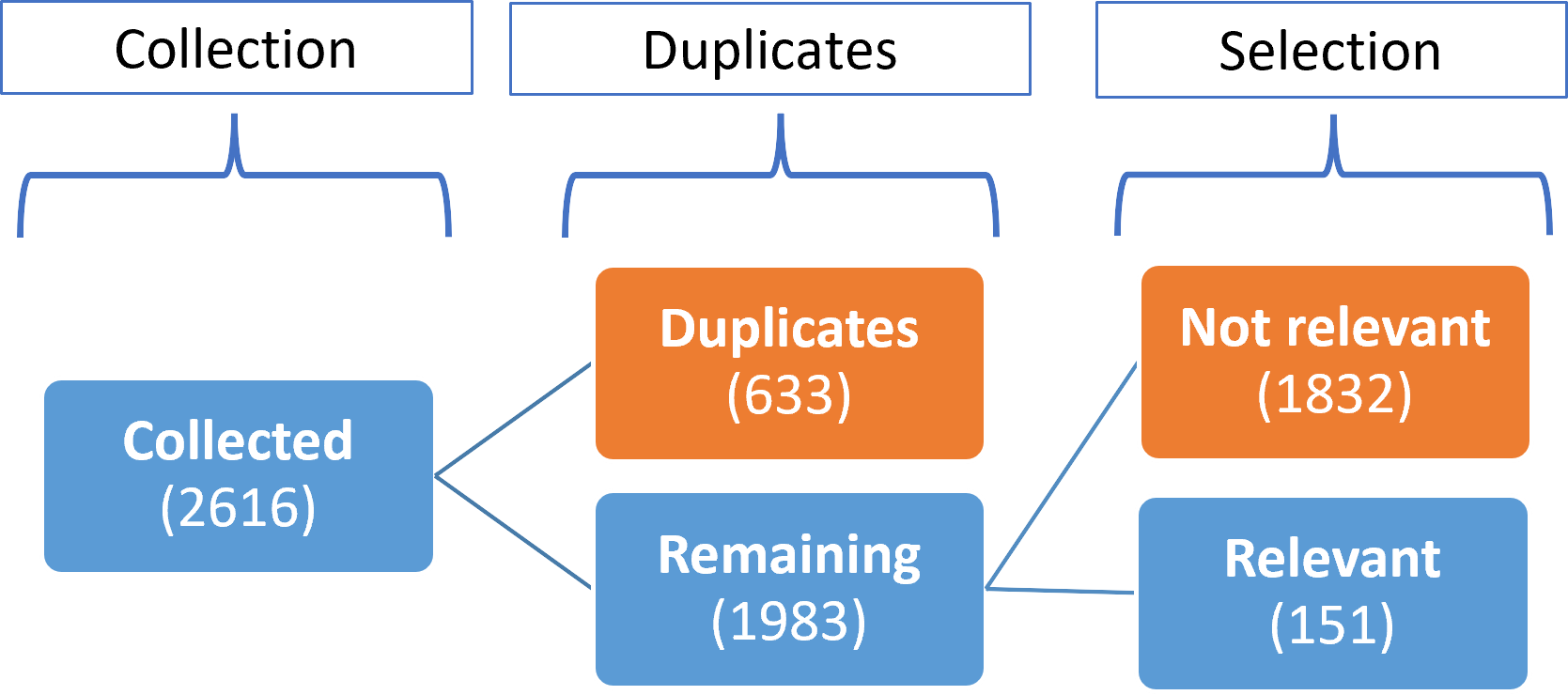}
            \caption{Articles collected at each phase of the systematic search process, including acquisition from various sources, removal of duplicates, and study selection, which also involved quality assessment.}
            \label{fig:selection}
        \end{figure}

        The results of study collection and duplicate removal are illustrated in Figure \ref{fig:selection}. A total of 2,616 studies were collected from the different sources using the aforementioned queries, of which 633 duplicates were detected and removed, leaving a total of 1,983 studies.

        
    \subsection{Study selection} \label{section:selection}
    

        After collecting studies from various sources, limiting by year, and removing duplicates, exclusion criteria were applied to eliminate non-relevant studies. The exclusion criteria were as follows:
        
        \begin{itemize}


            \item \textbf{Deep Learning}: Studies not utilizing DL were considered irrelevant for this review. Including this criterion in the exclusion criteria rather than in the query strings enabled the inclusion of more relevant studies, since many studies did not directly reference DL but instead used the name of a specific model.
        

            \item \textbf{Language}: Studies not in English or Spanish were excluded.
            
            
            \item \textbf{Data Type}: Studies using data types other than RGB, depth, or IR were excluded. This includes both videos and images. Skeleton data was also included, but only if computed from the other three types of data. Studies using sensory data along with visual data were also included, allowing for multimodal approaches.
            
            
            \item \textbf{Accessibility}: Studies not accessible for various reasons, such as being part of paid content (e.g., book chapters), source website down, or retracted content, were excluded.
            
            
            \item \textbf{Redundancy}: In cases where a journal article extended a work already presented in a conference, the conference proceedings publications were omitted, as the journal article represented an extension of the same work.
            
            
            \item \textbf{Task}: Studies focused on tasks other than HAR or fall detection, such as velocity estimation, gait trend, level of tiredness, etc., were excluded. However, studies that did not directly perform HAR or fall detection but presented a new dataset for these tasks were included.
            
            
            \item \textbf{Target Collective}: Studies not centered on elderly people were excluded. Merely mentioning the elderly as one of the beneficiaries of the work was insufficient; the study had to either use data from elderly people or have them in mind when designing the experiment.
            
            
            \item \textbf{Works in Progress}: Conference proceedings about works in progress, containing only the initial stages of the study and lacking the experimentation phase, were excluded.

            
            \item \textbf{Quality}: Publications with very poor quality (e.g., null reproducibility, highly biased decisions, too small datasets, etc.) were excluded. More information about quality assessment can be found in Section \ref{section:quality}.

        \end{itemize}


    As depicted in Figure \ref{fig:selection}, only 151 studies remained after applying the exclusion criteria, comprising 89 conference proceedings and 64 journal articles. The conference proceedings were retained for analysis among the relevant studies, as they serve as a standard search strategy to address publication bias, which can lead to systematic bias in systematic reviews unless special efforts are made to address this issue \cite{slr-software}.
    
    \subsection{Quality assessment} \label{section:quality}
    

        Given the absence of a universally agreed-upon definition of study "quality," the proposed guidelines in \cite{slr-software} were adhered to, primarily focusing on bias and validity as measures of quality. Specifically, the following aspects were taken into account:

        \begin{itemize}
        

            \item \textbf{Reproducibility}: Assessing whether the work can be replicated. This can be achieved by disclosing the dataset used, using external datasets, and either publishing the code used for the model or providing sufficient details to recreate the model.
            

            \item \textbf{Comparison with Other Works}: Evaluating whether the performance of the model is compared with the state-of-the-art. It's essential to ensure that comparisons are made under fair conditions, meaning that the models should be trained and tested on the same data to avoid introducing bias.
            

            \item \textbf{Use of External Datasets}: Considering whether the model is tested on external datasets to mitigate possible bias from the data and facilitate comparison with other models for the same task. Additionally, using external datasets allows other studies to utilize the results without the need to retrain the model on different data.
        
        \end{itemize}


        These aspects were included in the list of fields during the data extraction phase (last three fields), as discussed in Section \ref{section:extraction}. Moreover, these quality aspects were also used as exclusion criteria, as previously mentioned in Section \ref{section:selection}.


        In addition to these aspects, the type of study, either conference proceedings or journal articles, was also considered as a quality indicator, with journal articles typically being longer and more mature.
    
    \subsection{Data extraction and synthesis} \label{section:extraction}
    

        From each study remaining after applying the exclusion criteria, various data points were extracted to summarize the content and establish taxonomies for various aspects of interest. All data were compiled into a table, with each entry containing the following fields:
        
       \begin{itemize}
            \itemsep0em 
            \item Title
            \item Author/s
            \item Type (journal article or conference proceedings)
            \item Publication year
            \item Task (HAR or fall detection)
            \item Data type (RGB, depth, IR or skeleton)
            \item Auxiliary sensor data type (accelerometers, gyroscopes, etc.)
            \item Camera used
            \item Dataset (name of external dataset/s or "custom")
            \item DL model/s and task (skeleton joints estimation, feature extraction, classification, etc.)
            \item Other ML models or computer vision techniques used
            \item System integration in a robot (yes/no) and which one
            \item System integration in a framework (yes/no)
            \item How is privacy preserved? (depth or IR only, low resolution, etc.)
            \item Reproducible (yes/no)
            \item Test with external datasets
            \item Comparison with other approaches
       \end{itemize}


        The complete list of relevant studies is provided in Tables \ref{tab:studies} and \ref{tab:studies2}, which display only basic information for each study. The remaining information will be synthesized in Section \ref{section:results} through tables and plots, allowing for an overview of the distribution of works by used data types, DL model families, datasets, etc. Additionally, particularly relevant or interesting aspects of the works will be summarized, and important concepts will be addressed in more detail.
        
    
\section{Results} \label{section:results}
    
        


    This section provides an overview of the primary studies discovered through the systematic search process and presents the findings. Each study is thoroughly examined, and summaries are presented in the form of tables and graphs where applicable. Subsections are structured to address individual research questions, enhancing readability and organization.

        \FloatBarrier
        \begin{table}[h]
            \label{tab:studies}
            \caption{Full list of relevant studies examined in this systematic review. The tasks of the studies may involve FD (fall detection), HAR (Human Action Recognition), or both (FD, HAR). The CV data column indicates the type of computer vision data used: RGB, D (depth), or IR (infrared). Data types are separated by commas if the study requires all of them or by slashes if only one is necessary. \textit{(continued on the next page)}}
            \begin{tabular*}{.91\textwidth}{llll|llll}
                \toprule
                \textbf{Ref.} & \textbf{Year} & \textbf{Task} & \textbf{CV Data} & \textbf{Ref.} & \textbf{Year} & \textbf{Task} & \textbf{CV Data} \\
                \midrule
\cite{sudasinghe2023vision} & 2023 & FD & D & \cite{zhang2022visual} & 2022 & FD & RGB \\
\cite{wang2023abnormal} & 2023 & FD & RGB & \cite{MUVIM} & 2022 & FD & RGB/D/IR \\
\cite{marshal2023image} & 2023 & FD & RGB & \cite{li2022fall} & 2022 & FD & RGB-D \\
\cite{eltahir2023deep} & 2023 & FD & RGB & \cite{inturi2022novel} & 2022 & FD \& HAR & RGB \\
\cite{ke2023empowering} & 2023 & FD & RGB & \cite{suarez2022afar} & 2022 & FD \& HAR & RGB \\
\cite{agrawal2023enhanced} & 2023 & FD & RGB & \cite{ARFD-Net} & 2022 & FD \& HAR & RGB \\
\cite{dakare2023fall} & 2023 & FD & RGB & \cite{rajavel2022iot} & 2022 & FD \& HAR & RGB \\
\cite{m2023fall} & 2023 & FD & RGB & \cite{wang2022lightweight} & 2022 & FD \& HAR & RGB \\
\cite{paul2023human} & 2023 & FD & RGB & \cite{xie2022privacy} & 2022 & FD \& HAR & RGB \\
\cite{fan2023intelligent} & 2023 & FD & RGB & \cite{patsch2022automatic} & 2022 & HAR & D \\
\cite{cheng2023research} & 2023 & FD & RGB & \cite{guerra2022neural} & 2022 & HAR & D \\
\cite{inturi2023synergistic} & 2023 & FD & RGB & \cite{sun2022real} & 2022 & HAR & D \\
\cite{wahla2023visual} & 2023 & FD & RGB & \cite{kim2022care} & 2022 & HAR & RGB \\
\cite{jain2023privacy} & 2023 & FD \& HAR & D & \cite{lin2022adaptive} & 2022 & HAR & RGB \\
\cite{rezaei2023unobtrusive} & 2023 & FD \& HAR & IR & \cite{he2022elderly} & 2022 & HAR & RGB \\
\cite{yazici2023smart} & 2023 & FD \& HAR & RGB & \cite{zhang2022bed} & 2022 & HAR & RGB \\
\cite{zhang2023deep} & 2023 & FD \& HAR & RGB & \cite{prasad2022deep} & 2022 & HAR & RGB \\
\cite{gao2023development} & 2023 & FD \& HAR & RGB & \cite{achirei2022human} & 2022 & HAR & RGB \\
\cite{gaya_morey2023explainable} & 2023 & FD \& HAR & RGB & \cite{islam2022multimodal} & 2022 & HAR & RGB \\
\cite{wang2023fall} & 2023 & FD \& HAR & RGB & \cite{isoi2022performance} & 2022 & HAR & RGB \\
\cite{rashid2023hac} & 2023 & FD \& HAR & RGB & \cite{ji2022design} & 2022 & HAR & RGB/D \\
\cite{luo2023human} & 2023 & FD \& HAR & RGB & \cite{ouyang2022cosmo} & 2022 & HAR & RGB-D \\
\cite{ravankar2023real} & 2023 & FD \& HAR & RGB & \cite{galvao2021framework} & 2021 & FD & RGB \\
\cite{singh2023real} & 2023 & FD \& HAR & RGB & \cite{galvão2021multimodal} & 2021 & FD & RGB \\
\cite{sukreep2023recognizing} & 2023 & FD \& HAR & RGB & \cite{kang2021study} & 2021 & FD & RGB \\
\cite{MSSkip} & 2023 & FD \& HAR & RGB & \cite{raj2021active} & 2021 & FD & RGB \\
\cite{zhou2023towards} & 2023 & FD \& HAR & RGB & \cite{yang2021edge} & 2021 & FD & RGB \\
\cite{zaghdoud2023metaplastic} & 2023 & HAR & RGB & \cite{sultana2021classification} & 2021 & FD & RGB \\
\cite{siow2023one} & 2023 & HAR & RGB & \cite{divya2021docker} & 2021 & FD & RGB \\
\cite{shejy2023activity} & 2023 & HAR & RGB & \cite{chen2021fall} & 2021 & FD & RGB \\
\cite{liu2023amir} & 2023 & HAR & RGB & \cite{killian2021fall} & 2021 & FD & RGB \\
\cite{HACER} & 2023 & HAR & RGB & \cite{ge2021human} & 2021 & FD & RGB \\
\cite{snoun2023deep} & 2023 & HAR & RGB/D & \cite{pita2021indoor} & 2021 & FD & RGB \\
\cite{EatSense} & 2023 & HAR & RGB-D & \cite{vaiyapuri2021internet} & 2021 & FD & RGB \\
\cite{ouyang2023harmony} & 2023 & HAR & RGB-D & \cite{liu2021privacy} & 2021 & FD & RGB \\
\cite{fayad2022elderly} & 2022 & FD & D & \cite{zherdev2021producing} & 2021 & FD & RGB \\
\cite{meraikhi2022multimodal} & 2022 & FD & RGB & \cite{feng2021research} & 2021 & FD & RGB \\
\cite{chen2022approach} & 2022 & FD & RGB & \cite{xie2021skeleton} & 2021 & FD & RGB \\
\cite{anwary2022deep} & 2022 & FD & RGB & \cite{fatima2021unsupervised} & 2021 & FD & RGB \\
\cite{lau2022fall} & 2022 & FD & RGB & \cite{berlin2021vision} & 2021 & FD & RGB \\
\cite{FallAction} & 2022 & FD & RGB & \cite{li2021fall} & 2021 & FD & RGB/D \\
\cite{aarthi2022intelligent} & 2022 & FD & RGB & \cite{fernando2021computer} & 2021 & FD \& HAR & RGB \\
\cite{VWFP} & 2022 & FD & RGB & \cite{tseng2021elder} & 2021 & FD \& HAR & RGB \\
\cite{galvão2022onefall} & 2022 & FD & RGB & \cite{ramirez2021fall} & 2021 & FD \& HAR & RGB \\
\cite{zheng2022realization} & 2022 & FD & RGB & \cite{wang2021fall} & 2021 & FD \& HAR & RGB \\
\cite{zahan2022sdfa} & 2022 & FD & RGB & \cite{hasib2021vision} & 2021 & FD \& HAR & RGB \\
\cite{rajalaxmi2022vision} & 2022 & FD & RGB & \cite{wang2021action} & 2021 & HAR & D \\
\cite{anitha2022vision} & 2022 & FD & RGB & \cite{budisteanu2021combining} & 2021 & HAR & D \\
                \bottomrule
                \end{tabular*}
        \end{table}
        \FloatBarrier

        \FloatBarrier
        \begin{table}[h]
            \caption{Full list of relevant studies examined in this systematic review. The tasks of the studies may involve FD (fall detection), HAR (Human Action Recognition), or both (FD, HAR). The CV data column indicates the type of computer vision data used: RGB, D (depth), or IR (infrared). Data types are separated by commas if the study requires all of them or by slashes if only one is necessary. \textit{(continuation)}}
            \label{tab:studies2}
            \begin{tabular*}{.84\textwidth}{llll|llll}
                \toprule
                \textbf{Ref.} & \textbf{Year} & \textbf{Task} & \textbf{CV Data} & \textbf{Ref.} & \textbf{Year} & \textbf{Task} & \textbf{CV Data} \\
                \midrule
\cite{tu2021feddl} & 2021 & HAR & D & \cite{jaouedi2020prediction} & 2020 & HAR & RGB \\
\cite{tianming2021multi} & 2021 & HAR & D & \cite{lang2020research} & 2020 & HAR & RGB \\
\cite{lumetzberger2021privacy} & 2021 & HAR & D & \cite{PRECIS_HAR} & 2020 & HAR & RGB-D \\
\cite{badarch2021ultra} & 2021 & HAR & IR & \cite{mathe2020human} & 2020 & HAR & RGB-D \\
\cite{giannakos2021study} & 2021 & HAR & RGB & \cite{rafferty2019thermal} & 2019 & FD & IR \\
\cite{awal2021action} & 2021 & HAR & RGB & \cite{wang2019fall} & 2019 & FD & RGB \\
\cite{sivakumar2021computer} & 2021 & HAR & RGB & \cite{wang2019novel} & 2019 & FD & RGB \\
\cite{iksan2021implementation} & 2021 & HAR & RGB & \cite{brieva2019intelligent} & 2019 & FD & RGB \\
\cite{nambissan2021variegated} & 2021 & HAR & RGB & \cite{hassan2019convolution} & 2019 & FD & RGB \\
\cite{tan2021using} & 2021 & HAR & RGB & \cite{mohamed2019convolutional} & 2019 & FD & RGB \\
\cite{byeon2021body} & 2021 & HAR & RGB/D & \cite{kumar2019elderly} & 2019 & FD & RGB \\
\cite{KIST_SynADL} & 2021 & HAR & RGB/D & \cite{FPDS} & 2019 & FD & RGB \\
\cite{han2020two} & 2020 & FD & RGB & \cite{ferooz2019person} & 2019 & FD & RGB \\
\cite{chiang2020deep} & 2020 & FD & RGB & \cite{safarzadeh2019real} & 2019 & FD & RGB \\
\cite{serpa2020evaluating} & 2020 & FD & RGB & \cite{huang2019video} & 2019 & FD & RGB \\
\cite{berardini2020fall} & 2020 & FD & RGB & \cite{ma2019fall} & 2019 & FD & RGB, IR \\
\cite{romaissa2020fall} & 2020 & FD & RGB & \cite{cameiro2019multi} & 2019 & FD & RGB-D \\
\cite{wang2020human} & 2020 & FD & RGB & \cite{siriwardhana2019classification} & 2019 & HAR & D \\
\cite{lv2020hybrid} & 2020 & FD & RGB & \cite{phyo2019deep} & 2019 & HAR & D \\
\cite{li2020multi} & 2020 & FD & RGB & \cite{saini2019kinect} & 2019 & HAR & D \\
\cite{chen2020vision} & 2020 & FD & RGB & \cite{TOYS} & 2019 & HAR & D \\
\cite{khraief2020elderly} & 2020 & FD & RGB-D & \cite{phyo2019complex} & 2019 & HAR & RGB \\
\cite{kharazian2020increasing} & 2020 & FD & RGB-D & \cite{nan2019human} & 2019 & HAR & RGB \\
\cite{tateno2020human} & 2020 & FD \& HAR & IR & \cite{mehr2019human} & 2019 & HAR & RGB \\
\cite{ALMOND} & 2020 & FD \& HAR & RGB & \cite{jalal2019multi} & 2019 & HAR & RGB \\
\cite{tan2020activity} & 2020 & HAR & RGB & \cite{ding2019rt} & 2019 & HAR & RGB \\
\cite{atikuzzaman2020human} & 2020 & HAR & RGB & \cite{priya2019temporal} & 2019 & HAR & RGB \\
\cite{gul2020patient} & 2020 & HAR & RGB &  &  &  &  \\
                \bottomrule
                \end{tabular*}
        \end{table}
        \FloatBarrier
    
    \subsection{RQ1: Fall detection and Human Activity Recognition} \label{section:rq1}

\begin{figure}
            \centering
            \includegraphics[width=0.4\textwidth]{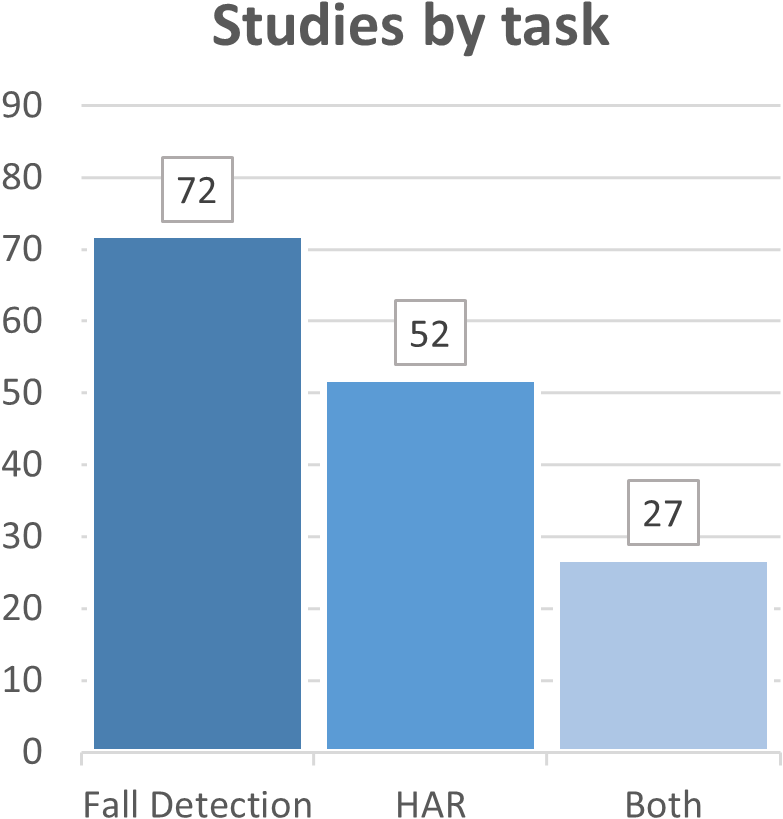}
            \caption{Distribution of studies by target task: fall detection, HAR or both.}
            \label{fig:tasks}
        \end{figure}

        The review primarily focuses on two main tasks: fall detection and Human Activity Recognition (HAR). It is worth noting that fall detection can be viewed an especially important activity of HAR. As illustrated in Figure \ref{fig:tasks}, fall detection has received the most attention in the past five years, with a total of 72 studies, while HAR has been explored in 52 studies. This discrepancy highlights the significance of fall detection when concerning the elderly population. Many works emphasize the importance of accurately and swiftly identifying falls among the elderly, given the potential for injuries and health implications if prompt actions are not taken. Consequently, several studies mention integrating fall detection into systems or applications capable of alerting medical personnel \cite{wahla2023visual, zhang2023deep}.


        Only 27 out of 151 studies (approximately 18\%) address both tasks simultaneously. This disparity arises from the emphasis placed on fall detection compared to other activities (such as walking or standing up), as well as the limited availability of data concerning fall scenarios, often resulting in an imbalanced problem. However, some studies manage to address both tasks. For instance, in \cite{inturi2022novel, ramirez2021fall, MSSkip}, both tasks are computed using the UP-FALL dataset \cite{UP-FALL}, which includes five types of falls and six common activities. This balanced dataset allows for the preservation of the importance of accurately detecting falls amidst other activities. A similar approach is adopted in \cite{wang2021fall}, where a custom dataset with egocentric videos is utilized. Nevertheless, there are studies that treat falls as just another task to recognize\cite{ALMOND, rajavel2022iot, yazici2023smart}.

    \subsection{RQ1.1: data type} \label{section:datatype}

\begin{figure}[h]
            \centering
            \includegraphics[width=0.4\textwidth]{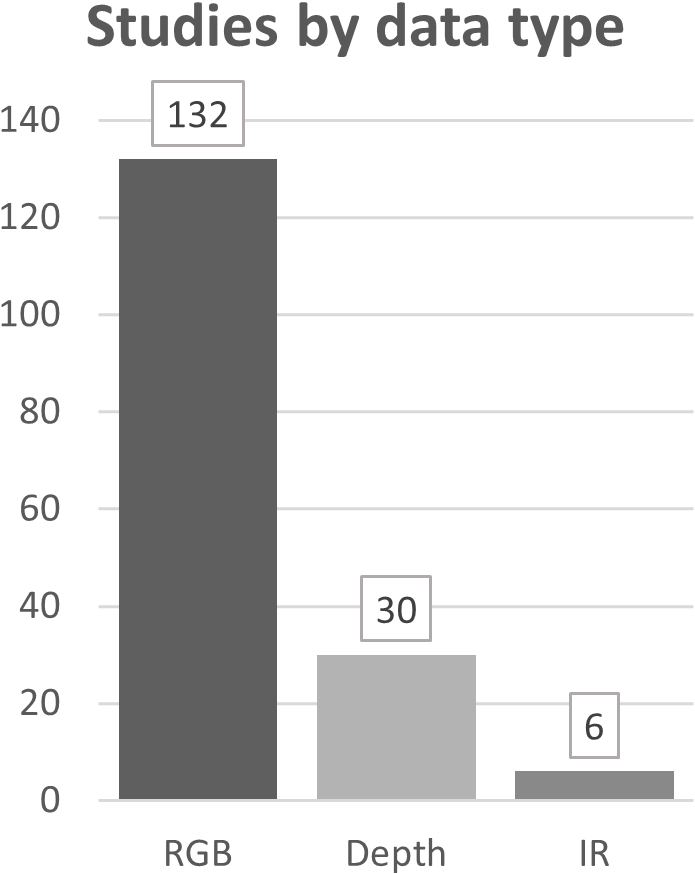}
            \caption{Distribution of studies by data type used. Note that more than one data type was used in some studies.}
            \label{fig:dataTypes}
        \end{figure}

        Among the studies collected, three types of vision data were considered: RGB, depth, and infrared (IR). The distribution of these data types is illustrated in Figure \ref{fig:dataTypes}. RGB data were the most prevalent for fall detection and HAR among the elderly (132 studies), followed by depth data (30 studies), with IR data being the least utilized (6 studies). This discrepancy can primarily be attributed to the accessibility of common cameras compared to specialized ones equipped with depth or infrared sensors. Additionally, RGB cameras offer benefits such as lower costs and easier visual data inspection. Notably, infrared cameras are less frequently employed, typically positioned overhead (top-down perspective) and characterized by very low resolutions, allowing for the use of simpler CNN models \cite{rezaei2023unobtrusive, rafferty2019thermal}, as well as non-convolutional models like LSTM \cite{badarch2021ultra, tateno2020human} and Transformer \cite{badarch2021ultra}. Depth cameras are more commonly used than infrared ones, although they are often employed to extract skeleton joints rather than directly performing fall detection and HAR. Specifically, 67\% of studies utilizing depth data computed skeleton joints before classification \cite{saini2019kinect, guerra2022neural, saini2019kinect}, while the remaining 33\% did not \cite{siriwardhana2019classification, khraief2020elderly, MUVIM}.

        Skeleton poses and sequences emerged as prevalent data types across the reviewed studies, with 67 studies incorporating skeleton data in some form. Given the human-centric nature of HAR and fall detection tasks, skeletal data represent logical features, offering efficient information compression while maintaining interpretability. Skeletons are typically represented as ordered sets of coordinates of body landmarks, either in 2D \cite{ravankar2023real, zahan2022sdfa, inturi2022novel} or 3D \cite{li2022fall, saini2019kinect, guerra2022neural} positions, depending on whether they were estimated from RGB or depth data, respectively. When skeleton estimation is performed on videos, the result is a sequence of skeleton poses with an added temporal dimension, enabling exploration of pose evolution over time intervals. 35 studies employed the evolution of one or more body landmarks for fall or HAR recognition \cite{zhang2023deep, zahan2022sdfa}, while the remaining 32 studies performed recognition using static poses exclusively \cite{kim2022care, PRECIS_HAR}.


        In addition to vision data, some studies utilized sensor data to enhance system performance, employing different models or strategies for classification and subsequently fusing the results. Fourteen studies, listed in Table \ref{tab:sensors}, utilized at least one of five types of sensor data, including Inertial Measurement Unit (IMU)\footnote{Studies not specifying the use of an IMU but using accelerometers and gyroscopes have also been included in this group.}, audio, barometer, luminosity, radar, electrocardiogram (ECG), GPS, and network traffic. IMU data was the most commonly used, featured in 10 of the 14 studies, particularly for fall detection (in 6 out of 10 studies using IMUs), owing to its effectiveness in identifying abrupt movements and subsequent immobility \cite{galvão2021multimodal, lv2020hybrid}. Barometer, GPS, radar, luminosity, and ECG data were consistently employed in conjunction with IMU data. Barometer and luminosity data served to acquire auxiliary or redundant information to enhance recognition consistency \cite{divya2021docker, lv2020hybrid}. ECG data in \cite{yazici2023smart} was utilized to identify inconsistencies in recognition and trigger specific further computations. In \cite{ouyang2023harmony}, four types of data (IMU, audio, radar, and GPS), along with visual data, were used for federated learning, where independent models were trained using different data modalities.

        \begin{table}
            \caption{Studies using multi-modal approaches and type of fusion with visual data.}
            \label{tab:sensors}
            \begin{tabular*}{0.75\textwidth}{lllll}
                \toprule
                \textbf{Ref.} & \textbf{Year} & \textbf{Task} & \textbf{Sensors} & \textbf{Fusion type}\\
                \midrule
\cite{ouyang2023harmony}              & 2023 & HAR     & IMU, Audio,   Radar, GPS & Intermediate \\
\cite{yazici2023smart}                & 2023 & HAR, FD & IMU, ECG                 & None         \\
\cite{liu2023amir}                    & 2023 & HAR     & Network traffic          & Intermediate \\
\cite{zhou2023towards}                & 2022 & HAR, FD & CSI                      & Late         \\
\cite{meraikhi2022multimodal}         & 2022 & FD      & IMU                      & Late         \\
\cite{lin2022adaptive}                & 2022 & HAR     & IMU                      & Late         \\
\cite{aarthi2022intelligent}          & 2022 & FD      & IMU                      & Intermediate \\
\cite{islam2022multimodal}            & 2022 & HAR     & IMU                      & Intermediate \\
\cite{ouyang2022cosmo}                & 2022 & HAR     & IMU, Radar               & Intermediate \\
\cite{iksan2021implementation}        & 2021 & HAR     & Audio                    & None         \\
\cite{galvão2021multimodal}           & 2021 & FD      & IMU                      & Intermediate \\
\cite{divya2021docker}                & 2021 & FD      & IMU, Luminosity          & Late         \\
\cite{lv2020hybrid}                   & 2020 & FD      & IMU, Barometer           & Late         \\
\cite{siriwardhana2019classification} & 2019 & HAR     & Audio                    & Intermediate \\
                \bottomrule
            \end{tabular*}
        \end{table}
        

        Regarding data fusion, no instances of early fusion were found. Instead, intermediate (7 studies) and late (5 studies) fusion methods were prevalent. Late fusion involved using a model for each data modality to produce a classification result, with the final classification determined using either voting \cite{meraikhi2022multimodal, divya2021docker} or weight attribution methods \cite{lin2022adaptive, lv2020hybrid, zhou2023towards}. In intermediate fusion, different models extracted features from various modalities, with a final model performing classification based on concatenated feature inputs. Various final models were utilized, including CNN \cite{galvão2021multimodal}, fully connected layers \cite{sultana2021classification, islam2022multimodal}, SVM \cite{aarthi2022intelligent}, and stacked classifiers \cite{liu2023amir}. In two studies, no fusion was performed, with different options provided for classification using distinct data modalities \cite{yazici2023smart, iksan2021implementation}.

    \subsection{RQ1.2: DL models} \label{section:models}

\begin{figure*}[h]
             \centering
             \includegraphics[width=.8\textwidth]{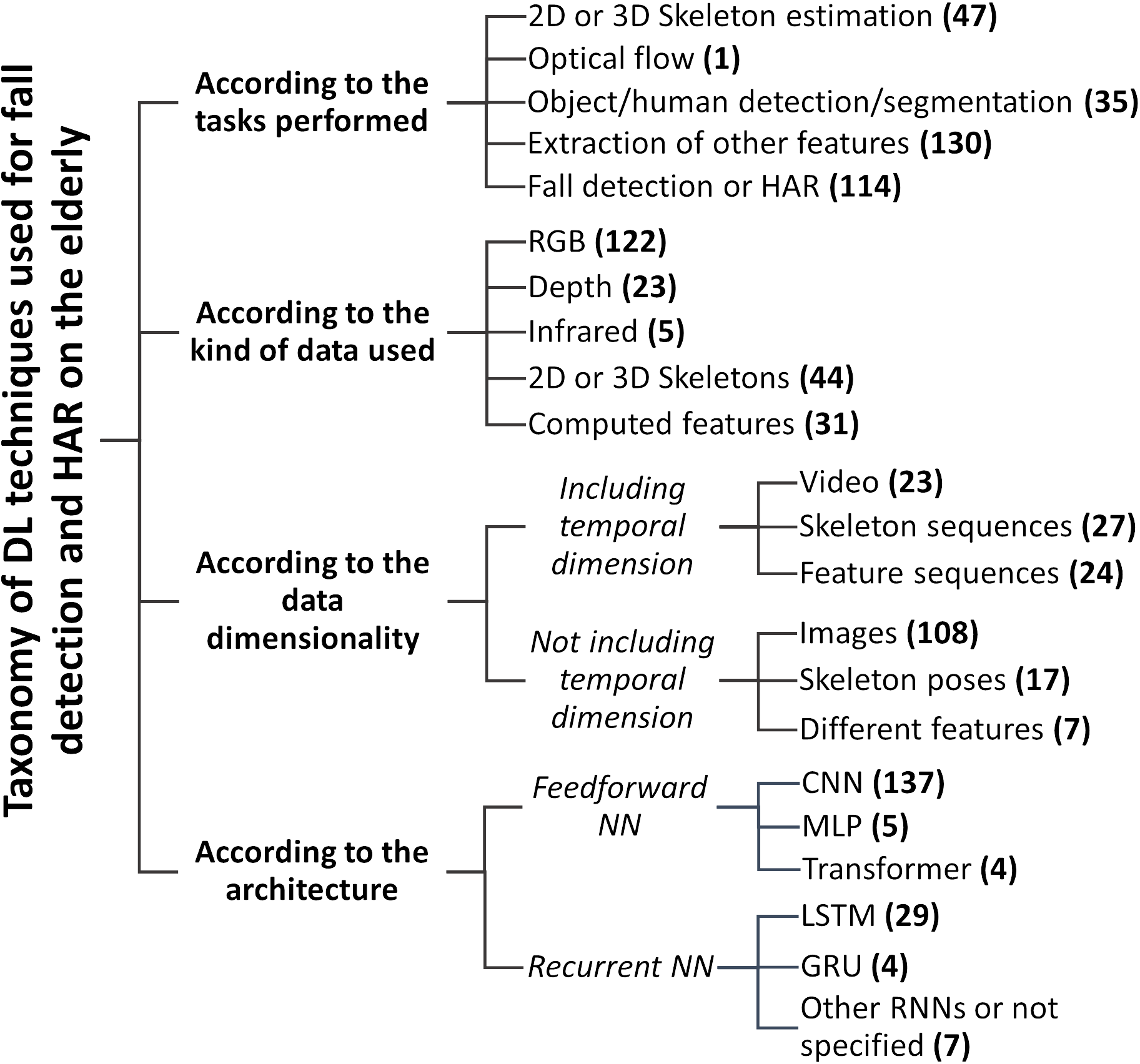}
             \caption{Taxonomy of the DL techniques used in the found studies. The number of studies where each category was used is displayed in bold. Note that multiple models were used in many studies, and hence the same study can be counted in more than one category.}
            \label{fig:taxonomy}
        \end{figure*}

        Table \ref{tab:models} provides a summary of all DL models utilized in the analyzed studies. These models are often employed for various specific tasks, including skeleton joints estimation, optical flow computation, and feature extraction. Moreover, the input data for these models encompasses not only images or videos but also features frequently computed by other DL models, such as 2D or 3D skeleton poses and optical flow. A taxonomy of the identified DL models, based on different characteristics, is presented in Figure \ref{fig:taxonomy}, offering a total count for each category. There is considerable diversity in the utilization of these models, regarding datasets used for evaluation, data types, and methodology. As demonstrated in the next section, in Table \ref{tab:datasets}, a wide range of datasets was employed across the analyzed studies, with many utilizing custom datasets. Additionally, prominent datasets like URFD and UP-FALL offer various data types, including RGB recordings, depth, skeletons, accelerometers, etc., which may lead to data differences even when studies are evaluated on the same dataset. The methodology for training and testing DL methods also varies across studies, with some employing k-fold cross-validation, leave-one-out cross-validation, or no cross-validation at all. Consequently, due to the lack of standardized conditions for a fair comparison, quantitative metric results were not included in Table \ref{tab:models}.



        \FloatBarrier
        \begin{table}[h]
            \caption{DL models utilized in the reviewed studies, tasks they are employed for, input data they process, and number of studies in which they are featured. Abbreviations are used for Fall Detection (FD), Object Detection (OD), and Object Segmentation (OS). Models of to the same family are grouped together, including R-CNN, LSTM, YOLO, VGG, ResNet, RNN, CNN, and GCN.}
            \label{tab:models}
            \begin{tabular*}{\textwidth}{ll|lll}
                \toprule
                    \multicolumn{2}{c|}{\textbf{Model}} & \multicolumn{3}{c}{\textbf{Studies}}\\
                    \midrule
                    \textbf{Ref.} & \textbf{Name} & \textbf{Task} & \textbf{Input data} & \textbf{N} \\
                    \midrule
\cite{OpenPose}       & OpenPose       & 2D Skeleton                            & RGB Image                             & 25      \\
\cite{AlphaPose}      & AlphaPose      & 2D   Skeleton                          & RGB   Image                           & 10      \\
\cite{MediaPipe}      & MediaPipe      & 2D Skeleton                            & RGB Image                             & 4       \\
\cite{PoseNet}        & PoseNet        & 2D   Skeleton                          & RGB   Image                           & 3       \\
\cite{MoveNet}        & MoveNet        & 2D Skeleton                            & RGB Image                             & 2       \\
\cite{RMPE}           & RMPE           & 2D   Skeleton                          & RGB   Image                           & 1       \\
\cite{PoseFlow}       & PoseFlow       & 2D Skeleton                            & RGB Image                             & 1       \\
                      & Baidu AI       & 2D   Skeleton                          & RGB   Image                           & 1       \\
\cite{FastPose}       & FastPose       & 2D Skeleton                            & RGB Image                             & 1       \\
\cite{MobileNet}      & MobileNet      & 2D   Skeleton, HAR, FD                 & RGB   Image                           & 7       \\
\cite{DeepHAR}        & DeepHAR        & 2D Skeleton, HAR, FD                   & RGB Image                             & 1       \\
\cite{PoseConv3D}     & PoseConv3D     & 3D   Skeleton                          & Depth   Image                         & 1       \\
\cite{STN}            & STN            & 3D Skeleton                            & RGB-D Image                           & 1       \\
\cite{AE}             & Autoencoder    & FD                                     & Different   features                  & 3       \\
\cite{GAN}            & GAN            & FD                                     & Different features                    & 2       \\
\cite{Siamese_CNNs}   & Siamese CNNs   & FD                                     & RGB or   Optical Flow Video           & 1       \\
\cite{FallAction}     & FallNet        & FD                                     & RGB Video                             & 1       \\
\cite{DeepFall}       & DeepFall       & FD                                     & RGB,   Depth or IR Video              & 1       \\
\cite{Sep-TCN}        & Sep-TCN        & FD                                     & Skeleton sequence                     & 1       \\
\cite{DCF-Net}        & DCF-Net        & Features                               & RGB   Image                           & 1       \\
\cite{SqueezeNet}     & SqueezeNet     & Features                               & RGB Image                             & 1       \\
\cite{EfficientNet}   & EfficientNet   & Features                               & RGB   Image                           & 1       \\
\cite{C3D}            & C3D            & Features                               & RGB Video                             & 1       \\
\cite{R-CNN}          & R-CNN          & Features (OD, OS), FD                      & RGB   Image                           & 7       \\
\cite{YOLO}           & YOLO           & Features (OD), HAR, FD                 & RGB Image                             & 26      \\
\cite{MSSkip}         & MSSkip         & Features (OS)                          & RGB   Image                           & 1       \\
\cite{PointRend}      & PointRend      & Features (OS)                          & RGB Image                             & 1       \\
\cite{LiteFlowNet}    & LiteFlowNet    & Features (Optical Flow)                & RGB   Video                           & 1       \\
\cite{Slowfast}       & Slowfast       & Features, HAR                          & RGB Video                             & 3       \\
\cite{InceptionV3}    & InceptionV3    & Features,   HAR                        & RGB or   Depth Image                  & 1       \\
                      & CNN            & Features, HAR, FD                      & Image, Video or different   features  & 52      \\
\cite{LSTM}           & LSTM           & Features,   HAR, FD                    & Skel.   sequence or Video features    & 29      \\
\cite{VGG}            & VGG            & Features, HAR, FD                      & RGB or Depth Image                    & 13      \\
\cite{ResNet}         & ResNet         & Features,   HAR, FD                    & RGB or   Depth Image or Skel. Pose    & 12      \\
\cite{GCN}            & GCN            & Features, HAR, FD                      & Skel. sequence or Video   features    & 9       \\
                      & RNN            & Features,   HAR, FD                    & Skel.   sequence or Video features    & 7       \\
\cite{I3D}            & I3D            & Features, HAR, FD                      & RGB Video                             & 5       \\
\cite{GRU}            & GRU            & Features,   HAR, FD                    & Skel.   sequence or Video features    & 4       \\
\cite{Transformer}    & Transformer    & HAR                                    & IR Image (8x8)                        & 2       \\
\cite{TANet}          & TANet          & HAR                                    & RGB   Video                           & 2       \\
\cite{TPN}            & TPN            & HAR                                    & RGB Video                             & 2       \\
\cite{iCAN}           & iCAN           & HAR                                    & Bounding   Boxes sequence             & 1       \\
\cite{Xception}       & Xception       & HAR                                    & RGB Image                             & 1       \\
\cite{TSN}            & TSN            & HAR                                    & RGB   Video                           & 1       \\
\cite{VST}            & VST            & HAR                                    & RGB Video                             & 1       \\
\cite{TimeSformer}    & TimeSformer    & HAR                                    & RGB   Video                           & 1       \\
\cite{Glimpse_Clouds} & Glimpse Clouds & HAR                                    & Skeleton sequence                     & 1       \\
\cite{AIA}            & AIA            & HAR                                    & Video   features and Bounding Boxes   & 1       \\
                      & MLP            & HAR, FD                                & Different features                    & 5       \\
\cite{AlexNet}        & AlexNet        & HAR, FD                                & Feature   Image                       & 1       \\
\cite{ARFD-Net}       & ARFD-Net       & HAR, FD                                & Skeleton sequence                     & 1       \\
                    \bottomrule
                \end{tabular*}
        \end{table}
        \FloatBarrier

        As mentioned in section \ref{section:datatype}, many studies utilize skeleton joints as features for fall detection and HAR. To estimate these joints, various DL models are employed, with OpenPose \cite{OpenPose} and AlphaPose \cite{AlphaPose} being the most prevalent (appearing in 25 and 10 studies, respectively). OpenPose utilizes a non-parametric representation (referred to as Part Affinity Fields) to detect skeleton joints from all humans in the image simultaneously, while AlphaPose performs human detection first and then predicts the skeleton joints for each individual. Subsequently, multiple models are used for fall detection and HAR with these skeleton joints:


        \begin{itemize}
            \item \textbf{Recurrent Networks:} Long-Short Term Memory (LSTM) \cite{STN, anwary2022deep, lau2022fall} and Gated Recurrent Unit (GRU) \cite{lau2022fall, jaouedi2020prediction} are commonly used, with others grouped as RNN \cite{kang2021study, sun2022real}.
            \item \textbf{Graph-Based Network:} The Graph-Convolutional Network (GCN) was the only one found, which treats skeletons as graphs rather than sequences \cite{lin2022adaptive, budisteanu2021combining, zahan2022sdfa}. Additionally, graph-based networks have the potential to perform collective activity recognition by leveraging interactive relations \cite{reviewer}.
            \item \textbf{Convolutional Networks:} Various models like VGG architectures \cite{atikuzzaman2020human, iksan2021implementation, cameiro2019multi}, MobileNet \cite{kim2022care, serpa2020evaluating, snoun2023deep}, ResNet family \cite{serpa2020evaluating, snoun2023deep}, among others \cite{phyo2019complex, phyo2019deep, PRECIS_HAR}, are employed.
        \end{itemize}


        Only one DL model, LiteFlowNet, was used for optical flow estimation across the studies \cite{lv2020hybrid}. However, 11 additional studies utilized optical flow at some stage of the recognition pipeline through non-DL-based methods \cite{yang2021edge, brieva2019intelligent, prasad2022deep, khraief2020elderly, cameiro2019multi, isoi2022performance, ji2022design, rajalaxmi2022vision, berlin2021vision, EatSense, wang2023fall}.
        

        Object detection was a prevalent task in the reviewed studies (found in 35 studies), with models from two families: R-CNN \cite{R-CNN} and YOLO \cite{YOLO}. R-CNN involves a multi-step process including region proposal, feature extraction, object classification, bounding box regression, and non-maximum suppression. Conversely, YOLO focuses on real-time object detection with a single pass through the image. Both models received several ameliorations in later versions. These models were utilized for various purposes across the studies:

        \begin{itemize}
            \item Obtaining a sequence of bounding boxes from scene objects, which can serve as features in next steps \cite{he2022elderly, achirei2022human, aarthi2022intelligent}.
            \item Triggering computation of fall detection or HAR upon detection of human presence, saving computation time \cite{fernando2021computer, zheng2022realization}.
            \item Reducing data complexity by putting the focus on the target person \cite{ALMOND, tseng2021elder, agrawal2023enhanced}.
            \item Getting features from the humans in the scene, like height-to-width ratio, used for fall detection or HAR in further steps \cite{FPDS, killian2021fall, m2023fall}.
            \item Direct detection of falls or recognition of activities \cite{chiang2020deep, ke2023empowering, chen2021fall}. 
        \end{itemize}


        Additionally, object segmentation plays a crucial role in several studies. The most commonly used model is Mask R-CNN \cite{Mask_R-CNN}, which extends the capabilities of the R-CNN family to object segmentation. Another notable model is PointRend \cite{PointRend}, a neural network module that enhances the granularity of segmentation models by treating image segmentation as a rendering problem. Conversely, a novel model proposed in \cite{MSSkip} specifically addresses object segmentation as part of the processing pipeline for fall detection and post-fall classification, named MSSkip. MSSkip builds upon common ideas from other segmentation models but incorporates multi-scale skip connections and depth-wise separable convolutions in the decoder to minimize computation. Object segmentation serves various purposes in the reviewed studies: in \cite{galvão2022onefall}, averaged output masks are utilized as spatio-temporal features for further recognition steps; \cite{zherdev2021producing} performs direct classification into fall or not fall based on the segmentation of fallen individuals; segmentation masks are fed to a convolutional LSTM in \cite{MSSkip} and to a CNN followed by an LSTM in \cite{chen2020vision} to extract spatial and temporal features for fall detection; in \cite{hasib2021vision}, segmentation masks are input to different machine learning models to identify falls. Conversely, in \cite{wahla2023visual}, segmentation is used solely to anonymize images before feeding them to an autoencoder for fall detection.


        Moreover, alongside the aforementioned DL-computed features, other features are predominantly computed using convolutional models such as VGG-16, VGG-19 \cite{berardini2020fall, agrawal2023enhanced, chen2020vision}, ResNet \cite{wang2023fall, romaissa2020fall}, or InceptionV3 \cite{kharazian2020increasing}. Less frequently, non-DL-based features like Histograms of Oriented Gradients (HOG) \cite{kumar2019elderly, wang2023fall}, Local Binary Patterns (LBP) \cite{liu2021privacy, wang2023fall}, and Bag of Words (BoW) \cite{prasad2022deep, ferooz2019person} are also utilized. Following feature extraction, multiple DL models are employed for classification. However, at this stage, it is common to use non-deep machine learning models such as Support Vector Machine \cite{wang2021fall, FPDS, ji2022design, wang2023fall}, Random Forest \cite{wang2023fall, inturi2023synergistic}, Decision Tree \cite{wang2021fall, inturi2023synergistic}, and KNN \cite{wang2021fall}.


        Furthermore, fall detection is frequently approached as a normal/abnormal classification task in the reviewed studies, with normal activities modeled and falls treated as abnormal data. This involves performing feature extraction, either using pre-trained models to extract spatio-temporal features from video/images or utilizing estimated skeleton joints, followed by training a model to identify normal activities. Various approaches are employed for this task, such as utilizing an MPED-RNN network on skeletal data \cite{fatima2021unsupervised}, employing DeepFall on multiple data modalities (RGB, depth, and IR) \cite{MUVIM}, using autoencoders after obtaining spatio-temporal features from other networks \cite{ma2019fall, anitha2022vision, wahla2023visual}, and employing Generative Adversarial Networks (GANs) by utilizing the discriminator as the normal/abnormal classifier \cite{liu2021privacy, galvão2022onefall}.


        Finally, the choice of architecture in the analyzed studies often depends on the data dimensionality, with recurrent neural networks (RNNs) primarily used when considering the temporal dimension and feedforward neural networks (FFNNs) when not. RNNs are well-suited for problems involving sequential data due to their ability to remember input data using internal memory. As such, they are often employed for fall detection and activity recognition from skeleton sequences \cite{zhang2023deep, STN} and feature sequences computed frame-wise by CNNs \cite{inturi2022novel, siriwardhana2019classification, berardini2020fall}. While CNNs are commonly used for extracting visual features from images, transformers have also been utilized in the FFNNs category, particularly for tasks involving low-resolution images \cite{badarch2021ultra}, 3D skeleton data \cite{snoun2023deep}, and video by adapting Vision Transformer (ViT) \cite{dosovitskiy2021image} to video formats \cite{gaya_morey2023explainable, HACER}. Additionally, multilayer perceptrons (MLPs) are consistently employed for skeleton data \cite{guerra2022neural, safarzadeh2019real, xie2021skeleton, xie2022privacy} or visual features \cite{hasib2021vision}.
        
    \subsection{RQ1.3: Datasets}


        Table \ref{tab:datasets} provides a comprehensive list of datasets used in the reviewed studies for activity recognition and fall detection. Emphasizing the importance of reproducibility and comparability, only publicly available datasets are included, aiming to facilitate future research in the field. Each dataset is categorized based on several common characteristics:
        
        \begin{itemize}
        

            \item \textbf{Elderly:} Despite fall detection and activity recognition often targeting elderly individuals, only a small fraction of datasets (12\%) include samples from this demographic. This scarcity highlights the challenge of collecting real-life data from the elderly population, especially genuine fall incidents.
            

            \item \textbf{Falls:} The majority of datasets (58\%) include falls as a class, with 23\% specifically focusing on binary classification between fall and not fall activities, underscoring the significance of this task in eldercare.
            

            \item \textbf{Type:} Video data is predominant (85\% of datasets), aligning with the temporal nature of activities like falls, where temporal context is crucial for accurate recognition. Furthermore, video allows for the rapid acquisition of a large quantity of images in the form of frames, which can then be utilized by data-driven solutions, such as DL-based methods.

        
        \FloatBarrier
        \setlength{\tabcolsep}{3pt}\begin{table}[h]
            \caption{Comprehensive list of publicly available datasets used in the reviewed studies, along with their basic specifications. The columns "Eld." and "Cl." denote the presence of elderly people in the datasets and the number of classes, respectively. The "Studies" column indicates in which studies they appear, or the number of reviewed studies if it is greater than two.}
            \label{tab:datasets}
            \begin{tabular*}{1.05\textwidth}{llllllllll}
                \toprule
        			\textbf{Ref.} & \textbf{Dataset} & \textbf{Eld.} & \textbf{Falls} & \textbf{Type} & \textbf{Data types} & \textbf{Samples} & \textbf{Cl.} & \textbf{Studies} \\
        			\midrule
\cite{URFD} & URFD & No & Yes & Video & RGB-D, Skel., IMU & 140 & 2 & 40 \\
\cite{UP-FALL} & UP-FALL & No & Yes & Video & RGB,   IR, IMU & 561 & 11 & 17 \\
\cite{Le2i} & Le2i & No & Yes & Video & RGB & 191 & 2 & 16 \\
\cite{MultiCam} & MultiCam & No & Yes & Video & RGB & 192 & 2 & 16 \\
\cite{NTU60} & NTU   RGB+D & No & Yes & Video & RGB-D,   Skel. & 56880 & 60 & 14 \\
\cite{FDD-Adhikari} & FDD-Adhikari & No & No & Image & RGB-D & 21499   (frames) & 5 & 6 \\
\cite{MSRDailyActivity3D} & MSRDailyActivity3D & No & No & Video & RGB-D, Skel. & 320 & 16 & 4 \\
\cite{UTD-MHAD} & UTD-MHAD & No & No & Video & RGB-D,   Skel., IMU & 861 & 27 & 4 \\
\cite{CAD-60} & CAD-60 & No & No & Video & RGB-D & 720 & 12 & 3 \\
\cite{ETRI} & ETRIActivity3D & Yes & Yes & Video & RGB-D,   Skel. & 112620 & 55 & 3 \\
\cite{HMDB51} & HMDB51 & No & Yes & Video & RGB & 6766 & 51 & 3 \\
\cite{KTH} & KTH & No & No & Video & RGB & 2391 & 6 & 3 \\
\cite{PRECIS_HAR} & PRECIS HAR & No & Yes & Video & RGB-D & 800 & 16 & 3 \\
\cite{TOYS} & ToyotaSmartHome & Yes & No & Video & RGB-D,   Skel. & 16115 & 31 & 3 \\
\cite{CAD-120} & CAD-120 & No & No & Video & RGB-D & 1200 & 10 & \cite{patsch2022automatic,   phyo2019complex} \\
\cite{DMLSmartActions} & DMLSmartActions & No & Yes & Video & RGB-D,   Skel. & 932 & 12 & \cite{yazici2023smart,   mehr2019human} \\
\cite{FPDS} & FPDS & No & Yes & Image & RGB & 2064 & 2 & \cite{VWFP, FPDS} \\
\cite{HQFSD} & HQFSD & Yes & Yes & Video & RGB & 55 & 2 & \cite{he2022elderly,   li2020multi} \\
\cite{NTU120} & NTU RGB+D 120 & No & Yes & Video & RGB-D, Skel. & 114480 & 120 & \cite{KIST_SynADL, zahan2022sdfa} \\
\cite{N-UCLA} & N-UCLA & No & No & Video & RGB-D,   Skel. & 100 & 10 & \cite{ALMOND} \\
\cite{UCF101} & UCF101 & No & No & Video & RGB & 13320 & 101 & \cite{berardini2020fall, ji2022design} \\
\cite{UTKinect-Action3D} & UTKinect-Action3D & No & No & Video & RGB-D,   Skel. & 200 & 10 & \cite{phyo2019deep,   tseng2021elder} \\
\cite{UWA3DII} & UWA3DII & No & Yes & Video & RGB-D, Skel. & 120 & 30 & \cite{ALMOND, zahan2022sdfa} \\
\cite{MUVIM} & MUVIM & Yes & Yes & Video & RGB-D,   IR, IMU & 244 & 2 & \cite{MUVIM} \\
\cite{ALMOND} & ALMOND & No & Yes & Video & RGB & 7565 & 22 & \cite{ALMOND} \\
\cite{BIT-interaction} & BIT-interaction & No & No & Video & RGB & 400 & 8 & \cite{jalal2019multi} \\
\cite{C-MHAD} & C-MHAD & No & Yes & Video & RGB, IMU & 120 & 7 & \cite{lin2022adaptive} \\
\cite{FallAction} & FallAction & No & Yes & Video & RGB & 2000 & 20 & \cite{FallAction} \\
\cite{FDDZeyuChen} & FDD-Chen & No & Yes & Video & RGB & 30 & 2 & \cite{wang2019novel} \\
\cite{FDD-TST} & FDD-TST & No & Yes & Video & RGB-D,   Skel., IMU & 132 & 8 & \cite{kharazian2020increasing} \\
\cite{FPDS-Elderly} & FPDS-Elderly & Yes & Yes & Image & RGB & 413 & 2 & \cite{VWFP} \\
\cite{IXMAS} & IXMAS & No & No & Video & RGB,   MHV & 330 & 11 & \cite{ALMOND} \\
\cite{Kinetics_400} & Kinetics 400 & No & No & Video & RGB & 306245 & 400 & \cite{nambissan2021variegated} \\
\cite{Kinetics_600} & Kinetics   600 & No & Yes & Video & RGB & 495547 & 600 & \cite{FallAction} \\
\cite{Kinetics_700-2020} & Kinetics 700-2020 & No & Yes & Video & RGB & 647907 & 700 & \cite{sivakumar2021computer} \\
\cite{KIST_SynADL} & KIST   SynADL & No & Yes & Video & RGB-D,   Skel. & 462200 & 55 & \cite{KIST_SynADL} \\
\cite{MMU} & MMU & No & Yes & Video & RGB & 51 & 2 & \cite{anwary2022deep} \\
\cite{NAD} & NAD & No & No & Video & RGB & 84 & 7 & \cite{KIST_SynADL} \\
\cite{OOPS} & OOPS & No & Yes & Video & RGB & 20338 & 2 & \cite{FallAction} \\
\cite{PKU-MMD} & PKU-MMD & No & Yes & Video & RGB-D,   Skel. & 21545 & 51 & \cite{giannakos2021study} \\
\cite{Stanford40} & Stanford40 & No & No & Image & RGB & 9532 & 40 & \cite{sivakumar2021computer} \\
\cite{V-COCO} & V-COCO & No & No & Image & RGB & 10346 & 25 & \cite{achirei2022human} \\
\cite{VWFP} & VWFP & No & Yes & Image & RGB & 6071 & 2 & \cite{VWFP} \\
\cite{YTBF} & YTBF & No & Yes & Video & RGB-D,   Skel. & 606 & 2 & \cite{li2020multi} \\
        			\bottomrule
                \end{tabular*}
        \end{table}
        \FloatBarrier
            

            \item \textbf{Data types:} While RGB data is ubiquitous, depth frames, skeleton joints, and inertial data are found in 38\%, 29\%, and 13\% of datasets, respectively. Other data types such as infrared data and motion history volumes (MHV) are less common. The presence of RGB data in all datasets allows for the discovery of the exact conditions of the recordings (environment, perspective, users, etc.) and serves as a visual check of the data, a feature not offered by other types of data.
            

            \item \textbf{Samples:} Dataset sizes vary significantly, ranging from less than 50 samples (e.g., FDD-Chen) to over 500,000 samples (e.g., Kinetics 700-2020), reflecting the diversity in data availability.
            

            \item \textbf{Classes:} The number of classes also varies widely, from binary classification to datasets with hundreds of classes, though the latter are typically not focused on AAL.
            
            
            \item \textbf{Studies:} Half of the datasets are utilized in only one study, while only five are used in more than ten studies, indicating varying degrees of dataset popularity and usage.
            
        \end{itemize}

        The University of Rzeszow Fall Detection (URFD) dataset \cite{URFD} stands out as the most extensively used, featuring in 40 studies \cite{wahla2023visual, chen2020vision, meraikhi2022multimodal}. Focused on fall detection, URFD offers 70 sequences capturing falls and activities of daily living (ADL) from two perspectives, along with various data modalities including RGB, depth, skeleton joints, and inertial data. The UP-FALL dataset \cite{UP-FALL}, appearing in 17 studies \cite{inturi2023synergistic, galvão2022onefall, inturi2022novel}, provides data from 17 subjects performing 11 activities, offering RGB video, infrared images, and inertial data for both fall detection and human activity recognition (HAR). In contrast, the Le2i dataset \cite{Le2i}, used in 16 studies \cite{yazici2023smart, han2020two, anwary2022deep}, focuses solely on fall detection, featuring 143 videos with falls and 48 with normal activities, with varying actors, scenery characteristics, and illumination conditions. Similarly, the MultiCam dataset \cite{MultiCam}, utilized in 16 studies \cite{agrawal2023enhanced, rajavel2022iot, sultana2021classification}, provides RGB video from 24 sequences captured from eight perspectives, facilitating the study of falls and confounding events. The NTU RGB+D dataset \cite{NTU60}, used in 14 studies \cite{budisteanu2021combining, tan2021using, PRECIS_HAR}, offers a vast collection of samples from 40 subjects performing 60 activities, recorded using Kinect cameras, thus providing RGB video, depth images, and skeleton joints. An extended version of this dataset also exists: the NTU RGB+D 120 dataset \cite{NTU120}, which expands upon it by adding 60 additional classes. However, it is only utilized in two of the reviewed studies \cite{KIST_SynADL, zahan2022sdfa}. The remaining datasets were utilized fewer than 10 times, with approximately half of them being employed in only one study.


        While most datasets are collected from real environments, two exceptions are noted: \cite{VWFP} and \cite{KIST_SynADL}, offering synthetic images and videos, respectively. Despite the advantages of synthetic data, such as ease of acquisition and controlled conditions, models trained solely on synthetic data may lack adaptability to real-world scenarios.

\begin{figure}[h]
            \centering
            \includegraphics[width=0.4\textwidth]{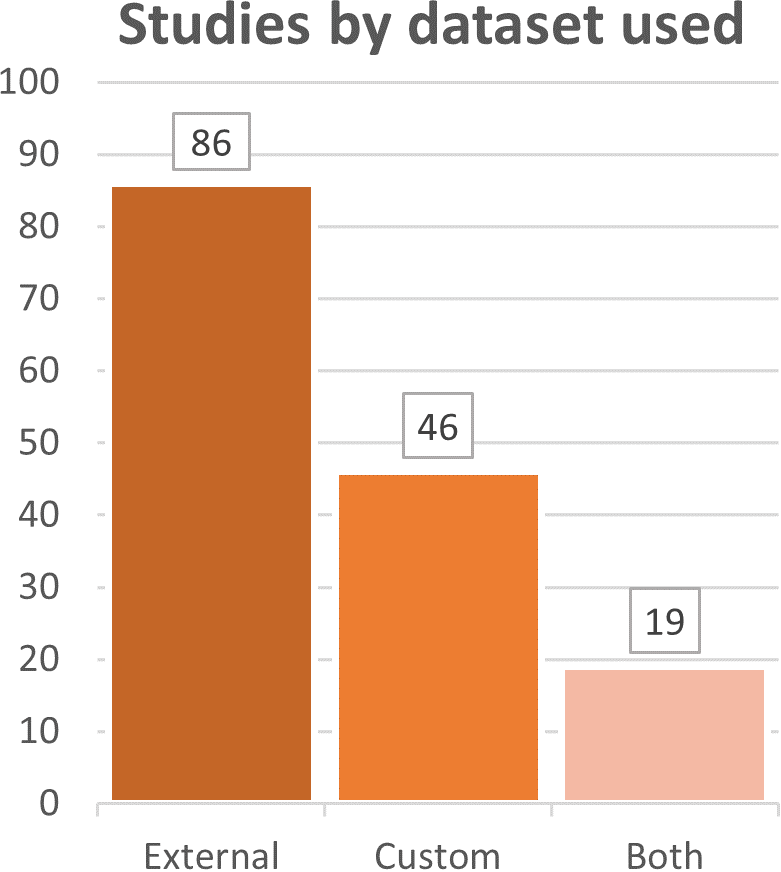}
            \caption{Distribution of studies by dataset used.}
            \label{fig:datasets}
        \end{figure}

        Notably, some studies opted for custom datasets instead of utilizing existing ones. Figure \ref{fig:datasets} illustrates the proportion of studies using custom, external, or both types of datasets. Only 19 studies provided evaluations on both custom and external datasets, with a greater frequency of evaluations conducted solely on external datasets (86 studies) compared to those exclusively using custom datasets (46 studies).
        
    \subsection{RQ2: Framework integration} \label{section:rq2}
    

        In 18 of the reviewed articles, frameworks were proposed to integrate the tasks of HAR or fall detection into real environments, addressing various aspects such as security, utilization of cloud services, client-server configuration, network communications, IoT devices, etc. Below, we provide brief descriptions of the proposed frameworks.

        In \cite{kim2022care}, a custom robot is suggested to integrate the HAR task into the environment, alongside other functionalities like language processing to enable chatbot interactions. In \cite{ALMOND}, a camera system is employed to capture visual data, which is then sent to a central server for computation. Subsequently, notifications, reports, and alerts are dispatched to a designated "guardian".

        In \cite{divya2021docker}, a Docker-based system is proposed to manage the flow between various programs involved in fall detection, distributing resources, and regulating communications. Docker is also utilized in \cite{killian2021fall}, where the NAO robot is suggested for data acquisition and user interaction to prevent falls. In \cite{rajavel2022iot, wang2022lightweight}, an intermediary step between recording and DL computation is introduced to preprocess video data and reduce bandwidth consumption.
        
        In \cite{ji2022design, zhang2022visual, zhang2023deep, paul2023human, achirei2022human, anwary2022deep, zheng2022realization}, the proposed frameworks integrate the collection of visual data through camera monitoring systems, centralized server-based recognition of fall detection or various activities, and trigger various responses based on the severity of the situation, such as contacting health services. For instance, \cite{paul2023human} utilizes the third-party service 'Twilio' to send phone messages in case of a fall, while in \cite{zheng2022realization}, the system transfers recordings to a computer for human inspection upon fall detection.
        
        In \cite{awal2021action, iksan2021implementation}, activity recognition results, along with recorded video data, are transmitted to a mobile application used for monitoring system users. Similar capabilities are offered in \cite{singh2023real}, with the addition of face blurring anonymization. \cite{liu2023amir} conducts all experiments in a connected environment, exploring the use of network traffic from multiple smart appliances combined with visual data to recognize various activities. Additionally, to assess the transferability of their approach across environments, they experimented with a smart residential apartment.
        
        In \cite{ouyang2023harmony}, federated learning is employed to ensure privacy preservation of users. The system incorporates three sensor modalities (depth, mmWave radar, and audio) and was tested in the homes of 16 elderly subjects.

    \subsection{RQ2.1: Hardware}


        A list of the hardware used in the reviewed studies (when mentioned) is presented in Table \ref{tab:hardware}. Specialized cameras such as thermal, depth, and wearable cameras, as well as social assistive robots, were included. Information regarding datasets not created in the reviewed studies was excluded. Hardware related to computation or common RGB cameras was omitted due to the wide range of possibilities available in these areas.

        \begin{table}[h]
            \caption{Special cameras and social robots found in the reviewed studies.}
            \label{tab:hardware}
            \begin{tabular*}{.82\textwidth}{llll}
                \toprule
                \textbf{Type} & \textbf{Model} & \textbf{Brand} & \textbf{Studies}\\
                \midrule
RGB-D Camera & Astra Pro & Orbbec & \cite{MUVIM,   lumetzberger2021privacy, PRECIS_HAR} \\
 & Kinect v1 & Microsoft & \cite{li2021fall, siriwardhana2019classification,   saini2019kinect, TOYS} \\
 & Kinect v2 & Microsoft & \cite{jain2023privacy,   sudasinghe2023vision, guerra2022neural} \\
 & RealSense D415 & Intel & \cite{EatSense} \\
                \midrule
Thermal Camera & AMG8831 & Panasonic & \cite{badarch2021ultra} \\
 & FLIR ONE & iTherml & \cite{MUVIM} \\
 & HTPA32x32d & Heimann Sensor & \cite{rafferty2019thermal} \\
 & MLX90640 & Melexis & \cite{rezaei2023unobtrusive} \\
 & MLX90641 & Melexis & \cite{tateno2020human} \\
 & PI450 & Optris & \cite{ma2019fall} \\
                \midrule
Wearable RGB Camera & OnReal G1 & Fondi & \cite{wang2023fall} \\
                \midrule
LiDAR & Horizon LiDAR & Livox & \cite{tu2021feddl} \\
                \midrule
Social Assistive Robot & Dori & Custom made & \cite{kim2022care} \\
 & LOLA & Custom made & \cite{FPDS} \\
 & NAO & Aldebaran URG & \cite{killian2021fall} \\
 & Pepper & Aldebaran URG & \cite{lang2020research, nan2019human} \\
                \bottomrule
            \end{tabular*}
        \end{table}


        For depth video retrieval, the most commonly used camera is the Microsoft Kinect (7 studies), followed by the Orbbec Astra Pro (3 studies), and Intel RealSense (1 study). These cameras share similar specifications, offering RGB-D recording using an IR camera for the depth channel, which provides accurate depth estimation at short distances. Additionally, they enable reliable 3D skeleton joint estimation.


        There is less consensus in the use of thermal cameras, with multiple camera models employed. Consequently, there is considerable variation in the retrieved data, including differences in resolution, sensitivity to temperature, maximum and minimum effective distances, etc.


        Only five studies deployed HAR or fall detection in an AAL system using a social assistive robot. Among these, two studies utilized the Pepper robot, one employed the NAO robot, and the remaining studies used custom-made robots.

    \subsection{RQ2.2: Privacy protection} \label{section:privacy}

\begin{figure}[h]
            \centering
            \includegraphics[width=.6\textwidth]{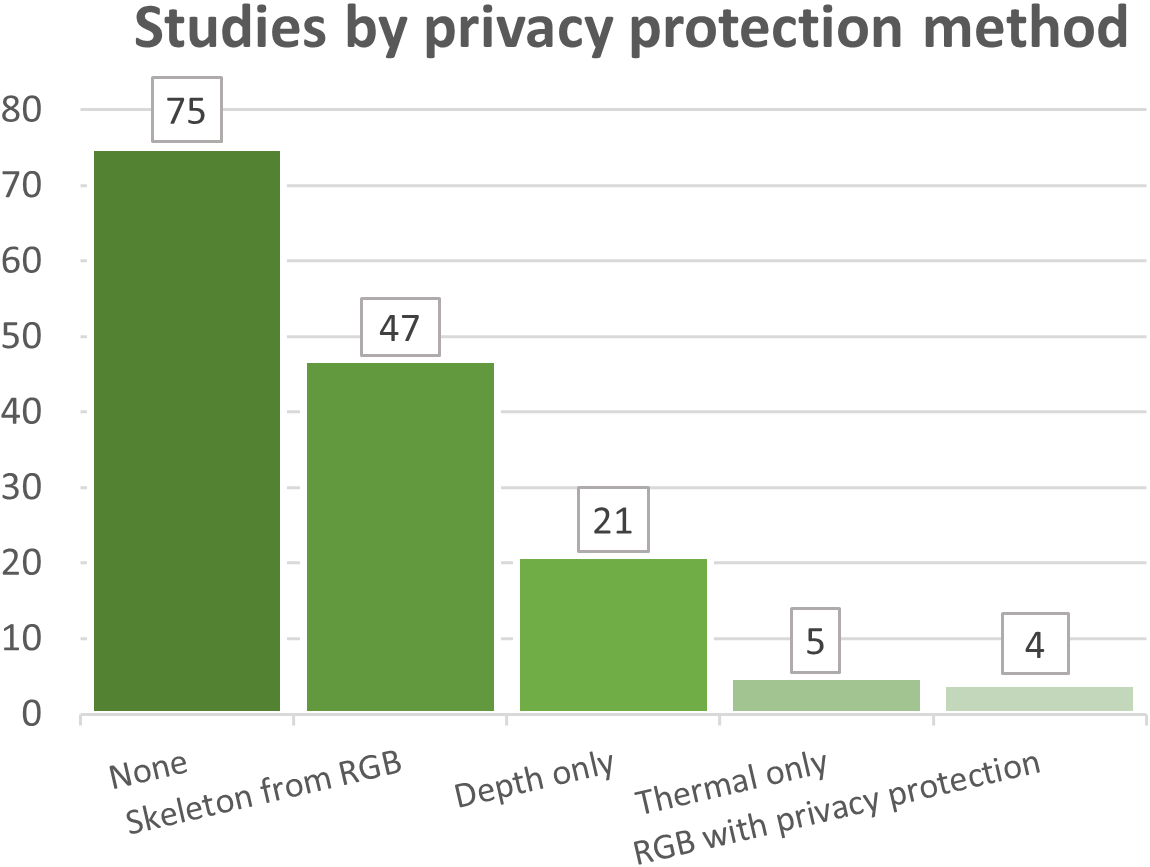}
            \caption{Distribution of studies by method used to preserve privacy. The total does not add up to 151 studies because in some studies different options were given.}
            \label{fig:privacy}
        \end{figure}

        Figure \ref{fig:privacy} illustrates the various privacy protection methods identified in the reviewed studies. Among the 151 studies reviewed, 75 did not address privacy concerns, opting for the use of unmodified RGB video or images of elderly users. Among the remaining studies, the majority employed skeleton data computed from RGB images, while four offered specific methods to anonymize RGB data, and others chose to utilize thermal or depth data instead.

        The most effective privacy-preserving methods avoid the deployment of RGB cameras in AAL settings. This is typically achieved through the use of visual data types that do not allow for subject identification, such as thermal and depth imaging. Among the collected studies, five exclusively employed thermal data \cite{tateno2020human, MUVIM, rafferty2019thermal, badarch2021ultra, rezaei2023unobtrusive}. In all cases, DL-based methods utilized CNNs to extract visual features and perform classification. Additionally, 21 studies utilized solely depth data, with 17 of them using it to estimate 3D skeleton poses, as demonstrated in \cite{saini2019kinect, guerra2022neural, snoun2023deep, jain2023privacy}. Notably, Microsoft Kinect was utilized in all 17 studies to estimate skeletons from depth maps through randomized decision forests \cite{kinect-skeletons}, leaving RGB data unused for this estimation. Four studies exploited depth data without skeleton estimation, instead relying on the extraction of human silhouettes \cite{siriwardhana2019classification} and visual features using CNNs \cite{lumetzberger2021privacy, MUVIM, tu2021feddl}.


        A total of 51 studies utilized RGB data at some stage, applying anonymization techniques. In contrast to the aforementioned studies, the input data used by these studies can be used to identify subjects, as conventional video recording is involved at the beginning of the processing pipeline. Among these, 47 studies relied on 2D skeleton estimation methods like OpenPose \cite{OpenPose} and AlphaPose \cite{AlphaPose} to protect privacy, removing visual data that can be used to identify users, as illustrated in \cite{inturi2022novel, han2020two, lin2022adaptive, ramirez2021fall, zahan2022sdfa}. There were four studies in which privacy was protected through other methods. In \cite{ma2019fall}, an IR camera is used to detect the face region of frames and remove it from the RGB frames. In \cite{liu2021privacy}, the RGB frames are modified in such a way that individuals cannot be identified, while fall detection can still be applied effectively. In \cite{wang2023fall}, a wearable camera providing a first-person perspective is used to avoid recording the user of the system. Human silhouettes are computed in \cite{wahla2023visual} and used for future recognition steps.
    
\section{Discussion}

\begin{figure*}[h]
         \centering
         \includegraphics[width=1\textwidth]{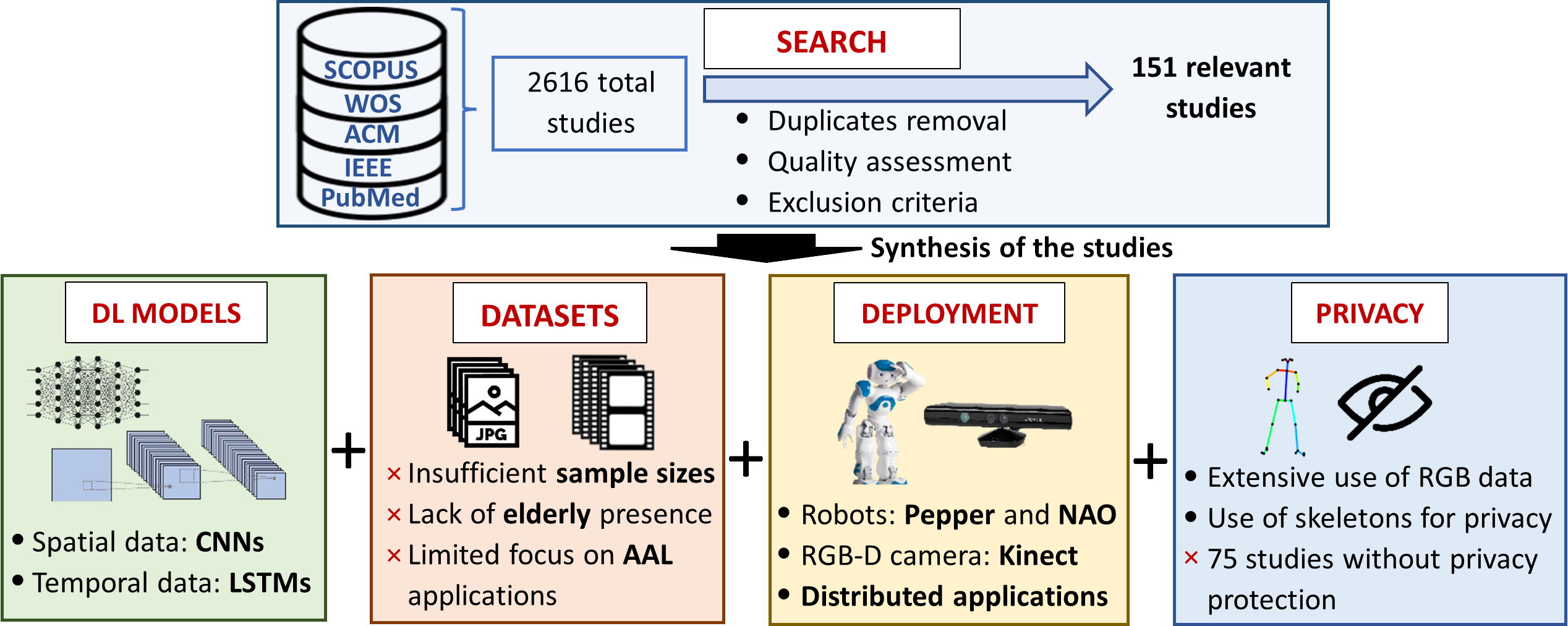}
         \caption{Summary of the search process and found results.}
        \label{fig:results}
    \end{figure*}

    This section utilizes the discovered results and the responses provided to the review questions to underscore common strengths and weaknesses of the reviewed studies. It also compiles a comprehensive list of recommendations for future reference based on the findings of this systematic literature review. In Figure \ref{fig:results}, the search process and key findings from the reviewed studies are summarized.

        \subsection{Strengths and weaknesses of the reviewed studies}
        

            Upon reviewing the 151 relevant studies and addressing the research questions, the main strengths and weaknesses observed are discussed in this subsection, which we believe can provide valuable insights for future studies in the field.


            A notable benefit of utilizing skeleton joints is their ability to significantly reduce data size compared to raw image or video data, while also offering user anonymization, maintaining data interpretability, and achieving satisfactory results in fall detection and HAR. Furthermore, there is a growing number of methods to derive human skeletons from RGB or depth data, with 13 different skeleton estimation DL models identified in the reviewed studies (as shown in Table \ref{tab:models}).


            The primary strength of studies employing only depth or infrared data lies in the privacy protection they afford, as RGB footage is not recorded at any point in the system pipeline. However, these studies also face two major weaknesses: a reduced amount of data for detection or recognition tasks, particularly pronounced in the case of IR recordings where resolution tends to be much lower, and less interpretable data, which may pose challenges when manual intervention is required to address errors.


            Among the reviewed studies, 27 perform both fall detection and HAR tasks (refer to Figure \ref{fig:tasks}). This integration is particularly significant, as it is often desirable to detect accidental falls while conducting HAR on elderly individuals. It is important to note that while fall detection can be integrated as another class during HAR, it should be computed separately due to its critical nature. Therefore, most studies including fall detection implement it differently than the recognition of other classes.


            Numerous studies have overlooked the temporal dimension when conducting HAR, thereby constraining the task significantly. This omission poses a significant weakness, particularly when incorporating activities that are challenging to distinguish without temporal data or are more effectively recognized with it, such as sitting/getting up or putting on/off clothes. Nonetheless, confining the analysis to spatial data typically offers the advantage of being faster and more straightforward.


            Regarding the choice of model architecture, convolutional models were found to be predominant. Their primary strengths lie in their effectiveness in processing spatial data and their extensive history, which has led to numerous improvements and architectural refinements across various fields and tasks. Given their suitability for image-based tasks, convolutional models are widely preferred and even have 3D versions tailored for video processing. In contrast, recurrent models excel in handling sequential data, thus complementing the capabilities of CNNs by facilitating the tracking of computed features across different frames. Multi-layer perceptrons, however, do not yield favorable results with spatial or sequential data; they are typically employed for classification based on computed features, akin to fully connected layers in a convolutional neural network. Transformer-based architectures, being relatively newer, are not as ubiquitous. Despite their promise in handling sequential and vision data, their large parameter count presents challenges in training and deploying them on low-specification systems. Nonetheless, they have showcased significant potential across various domains.


            Given that fall detection and HAR for the elderly aim to assist this population in AAL settings, studies offering frameworks for deploying systems in real environments are of particular interest. Eighteen studies, described in Section \ref{section:rq2}, fall into this category.


            Utilizing only an external dataset may impact the applicability of the technique to specific situations or environments, but it allows for the comparison of different methods on the same data. Conversely, relying solely on a custom dataset yields the opposite effects. The primary drawback associated with using only custom-made datasets is the external validity of the findings, as it becomes challenging to compare results with other studies, especially if the custom data is not disclosed. Including an evaluation on external datasets not only distinguishes studies from previous ones but also enables future studies to build on the obtained performance. While the majority of the reviewed studies evaluate on existing datasets for fall detection and HAR, 46 exclusively perform evaluations on new custom datasets (as depicted in Figure \ref{fig:datasets}), limiting the reliability of the results without comparisons with existing techniques or models. Conversely, 19 studies utilize both custom and external datasets, leveraging the strengths of each approach: specialization on custom data and comparison with other methodologies.

            From a data perspective, three common weaknesses are evident in the datasets utilized: the absence of elderly individuals, a limited number of samples, and the inclusion of numerous classes unrelated to activities of daily living (ADL), which may render them less suitable for fall detection and HAR among elderly populations. Primarily, the majority of datasets (88\%) lack elderly participants, presenting challenges during deployment as they represent the target users of the system but are not represented in the training data. In this regard, datasets such as ETRIActivity3D, ToyotaSmartHome, MUVIM, and FPDS-Elderly would be more suitable. Additionally, a limited number of samples may prove insufficient for DL models to generalize effectively. Three of the four most extensive datasets contain fewer than 200 samples, while the remaining dataset contains fewer than 600, with approximately half of the utilized datasets containing less than 1,000 samples. Instead, datasets such as ETRIActivity3D, NTU RGB+D (or NTU RGB+D 120), or ToyotaSmartHome, all offering more than 10,000 samples, would yield better generalization results. Lastly, datasets should be tailored to focus on ADL rather than general HAR to avoid unnecessary classes for monitoring elderly individuals in their daily lives. For instance, Kinetics (400, 600, or 700) or UCF101 would not be suitable for the considered tasks as they comprise videos collected from the internet, potentially containing irrelevant activities and cuts.

        \subsection{Recommendations for future works}
        

            Based on the results of this SLR, a series of particularly important considerations, in our understanding, should be taken into account when conducting new studies on the topic.


            First and foremost, it is crucial to assess user privacy. As observed, the approach to privacy protection will likely influence the type of data used, ranging from conventional RGB data to modified RGB, depth, IR, or skeleton data, which prevent user identification in the footage. Therefore, we recommend considering privacy protection as a fundamental aspect from the outset of the study.


            Selecting an appropriate DL model for fall detection and HAR requires consideration of the deployment conditions. For embedded systems or edge-deployments, such as in social robots or mobile applications, compact models are preferred, such as MobileNet or EfficientNet—well-known CNNs specifically tailored for such devices. These models can be augmented with recurrent models like LSTM to accommodate temporal data. Conversely, if model size is not a constraint, 3D CNNs like I3D, TPN, TANet, SlowFast, and C3D are suitable for video data, while GCN can be applied to skeleton data. Alternatively, Transformer-based architectures like TimeSformer or VST are also an option for processing video input data.


            For model evaluation, utilizing a publicly available dataset is essential to enable comparison with existing models or techniques. Prominent datasets for fall detection include URFD, UP-FALL, MultiCam, and Le2i, while for HAR, UP-FALL and NTU RGB+D are commonly used. However, we encourage the adoption of ETRIActivity3D or ToyotaSmartHome, which offer a more extensive collection of video samples and include elderly participants. Both datasets support HAR, with ETRIActivity3D additionally containing falls and providing multiple perspectives from elderly users, diverse classes (at least 30), and various data modalities, including RGB, depth, and skeleton joints.


            In cases where a custom dataset is provided, authors are encouraged to make it publicly available. This facilitates its use in future studies, either directly or by merging it with other datasets to form a larger dataset, enhances the reproducibility of experiments, and enables comparison with newer models or techniques. RGB-D cameras, such as Microsoft Kinect, Orbbec Astra Pro, and Intel RealSense, are recommended for collecting custom datasets as they facilitate experimentation with various types of data, with depth data offering privacy preservation capabilities.


            When deploying the system in a real environment, the most common approach, as indicated by the reviewed studies, involves establishing a camera setup within the environment. This setup records data and transmits it to a central server for processing. It is also the most cost-effective option, depending on factors such as camera type, resolution, and processing requirements. Alternatively, for those preferring to use an assistive robot, both NAO and Pepper robots are viable solutions. These commercial robots come equipped with cameras, speakers, microphones, and other necessary components, offering customizable options to adapt to different projects and environments.

\section{Conclusions}

    In this systematic literature review, we have investigated fall detection and human activity recognition for the elderly, with a particular focus on deep learning techniques applied to computer vision data. Our study aimed to address two primary research questions related to the implementation of DL methods for these tasks and their deployment in real-world environments, considering hardware and privacy concerns.
    
    Throughout the review process, we analyzed 151 relevant studies, providing a structured overview of the main findings to facilitate accessibility for practitioners and researchers. The findings offer valuable insights into the effective implementation of DL techniques for fall detection and HAR in elderly care, which are becoming increasingly important in the context of Ambient Assisted Living (AAL) systems.
    
    Privacy emerged as a common concern, with 50\% of the reviewed studies lacking any measures to address it. The most prevalent privacy protection method identified was the use of skeleton joints estimation, employed in 45\% of the studies.
    
    Convolutional DL models were found to be predominant, owing to their effectiveness in processing spatial data and extensive history of refinement. However, we observed a lack of consideration for the temporal dimension in many studies, which limits the recognition of some activities.
    
    Regarding datasets, we identified three common weaknesses: the absence of elderly individuals, a limited number of samples, and the inclusion of numerous irrelevant classes for the AAL systems. We recommend datasets such as ETRIActivity3D and ToyotaSmartHome, which offer extensive samples and include elderly participants.
    
    Moving forward, we emphasize the importance of privacy assessment from the outset of studies and recommend selecting appropriate DL models based on deployment conditions. Utilizing publicly available datasets for model evaluation is crucial, and authors are encouraged to make custom datasets publicly available to enhance reproducibility and facilitate future research.
    
    In terms of deployment, camera setups within the environment were the most common approach identified, offering cost-effectiveness and flexibility. Alternatively, assistive robots like NAO and Pepper provide customizable options for deployment in various projects and environments.
    
    Overall, this SLR provides a comprehensive overview of recent advancements in DL-based fall detection and HAR for the elderly, offering valuable insights for researchers, practitioners, and policymakers involved in developing and implementing AAL technologies.

\backmatter


        

\section*{Declarations}

    \bmhead{Funding}
    Grant PID2019-104829RA-I00 funded by MCIN/AEI/10.13-039/ 501100011033, project EXPLainable Artificial INtelligence systems for health and well-beING (EXPLAINING). Grant PID2022-136779OB-C32 funded by MCIN/AEI/ 10.13039/501100011033 and by ERDF A way of making Europe, project Playful Experiences with Interactive Social Agents and Robots (PLEISAR): Social Learning and Intergenerational Communication. F. X. Gaya-Morey was supported by an FPU scholarship from the Ministry of European Funds, University and Culture of the Government of the Balearic Islands.
        
    \bmhead{Competing interests}
    The authors have no competing interests to declare that are relevant to the content of this article.
    
    \bmhead{Authors' contributions}
    F.Xavier Gaya-Morey: Conceptualization, Methodology, Validation, Investigation, Writing - Original Draft, Writing - Review \& Editing Preparation, Visualization.

    Cristina Manresa-Yee: Conceptualization, Methodology, Writing - Review \& Editing Preparation, Supervision, Project administration, Funding acquisition.

    Jose M. Buades-Rubio: Conceptualization, Methodology, Writing - Review \& Editing Preparation, Supervision, Project administration, Funding acquisition.
    
    \bmhead{Ethical and informed consent for data used}
    This article does not contain any studies with human participants or animals performed by any of the authors.
    
    \bmhead{Data availability and access}
    All relevant data for the study can be found, structured and organized in the form of tables, throughout this document. No additional data were generated.

    \bmhead{Acknoledgments}
    This version of the article has been accepted for publication, after peer review but is not the Version of Record and does not reflect post-acceptance improvements, or any corrections. The Version of Record is available online at: http://dx.doi.org/10.1007/s10489-024-05645-1.









\bibliography{sn-bibliography}

\end{document}